\documentclass[10pt,twocolumn,letterpaper]{article}

\usepackage[pagenumbers]{cvpr} 
\usepackage[accsupp]{axessibility}
\usepackage{graphicx}
\usepackage{amsmath}
\usepackage{amssymb}
\usepackage{booktabs}

\usepackage{amsmath,amsfonts,bm}

\def\eqref#1{equation~\ref{#1}}

\def\1{\bm{1}}

\def\vh{{\bm{h}}}

\def\vl{{\bm{l}}}

\def\vn{{\bm{n}}}

\def\vv{{\bm{v}}}

\def\vx{{\bm{x}}}

\DeclareMathAlphabet{\mathsfit}{\encodingdefault}{\sfdefault}{m}{sl}
\SetMathAlphabet{\mathsfit}{bold}{\encodingdefault}{\sfdefault}{bx}{n}

\usepackage{epsfig}
\usepackage{graphicx}
\usepackage{amsmath}
\usepackage{amssymb}
\usepackage{booktabs}
\usepackage{graphicx}
\usepackage{diagbox}
\usepackage{rotating}
\makeatletter
\@namedef{ver@everyshi.sty}{}
\makeatother
\usepackage{tikz}

\usepackage{url}

\newcommand{\Tref}[1]{Table~\ref{#1}}
\newcommand{\eref}[1]{Eq.~(\ref{#1})}

\newcommand{\fref}[1]{Fig.~\ref{#1}}

\newcommand{\sref}[1]{Sec.~\ref{#1}}

\usepackage{xspace}
\makeatletter
\DeclareRobustCommand\onedot{\futurelet\@let@token\@onedot}
\def\@onedot{\ifx\@let@token.\else.\null\fi\xspace}
 
\def\ie{\emph{i.e}\onedot}

\def\etal{\emph{et al}\onedot}
\makeatother

\newcommand{\step}{\mathop{\rm step}}

\title{Neural Reflectance for Shape Recovery with Shadow Handling}

\author{Junxuan Li $ ^{1,2}$, \qquad Hongdong Li $ ^{1}$  \\
Australian National University $ ^1$ \qquad Data61, CSIRO $ ^2$\\
\texttt{\{junxuan.li, hongdong.li\}@anu.edu.au} 
}

\usepackage[pagebackref,breaklinks,colorlinks]{hyperref}

\usepackage[capitalize]{cleveref}
\crefname{section}{Sec.}{Secs.}
\Crefname{section}{Section}{Sections}
\Crefname{table}{Table}{Tables}
\crefname{table}{Tab.}{Tabs.}

\def\cvprPaperID{9865} 
\def\confName{CVPR}
\def\confYear{2022}

\begin{document}

\maketitle

\begin{abstract}
This paper aims at recovering the shape of a scene with unknown, non-Lambertian, and possibly spatially-varying  surface materials. When the shape of the object is highly complex and that shadows cast on the surface, the task becomes very challenging. To overcome these challenges, we propose a coordinate-based deep MLP (multilayer perceptron) to parameterize both the unknown 3D shape and the unknown reflectance at every surface point. This network is able to leverage the observed photometric variance and shadows on the surface, and recover both surface shape and general non-Lambertian reflectance.  We explicitly predict cast shadows, mitigating possible artifacts on these shadowing regions, leading to higher estimation accuracy. Our framework is entirely self-supervised, in the sense that it requires neither ground truth shape nor BRDF. Tests on real-world images demonstrate that our method outperform existing methods by a significant margin. Thanks to the small size of the MLP-net, our method is an order of magnitude faster than previous CNN-based methods.
\end{abstract}

\section{Introduction}

Recovering the 3D shape of a non-Lambertian object from its multiple photometric images taken by a fixed camera remains a challenging task. The diverse nature of real-world materials manifests a wide range of specularities on the surface, impeding traditional photometric methods~\cite{wu2010robust,ikehata2012robust,mukaigawa2007analysis,wu2010photometric}. Moreover, shadows commonly appear in non-convex objects occluding part of the object surface, hindering surface normal estimation.  Previous attempts to handle shadows often rely on a rather restrictive Lambertian assumption~\cite{chandraker2007shadowcuts}. The problem becomes much complicated if both specularities and shadows appear on the surface. 



With the recent advent of deep learning, tremendous progresses have been made in many computer vision problems, and there is no exception for photometric 3D reconstruction~\cite{santo2017deep, ikehata2018cnn, li2019learning, chen2020deep, yao2020gps, logothetis2021px}.  Current existing deep learning methods often tackle the problem in a supervised training manner. The underlying physics principle of image formation are not duly utilized.  In addition, the lack of interpretability of deep learning methods prevents leveraging the interactions between object appearance and surface normals. Despite various synthetic datasets with augmentation strategies~\cite{santo2017deep, ikehata2018cnn, chen2018ps, logothetis2021px}, it remains an open challenge to process real-world images with both specularities and shadows.

\begin{figure*}
\centering
\includegraphics[width=0.95\textwidth]{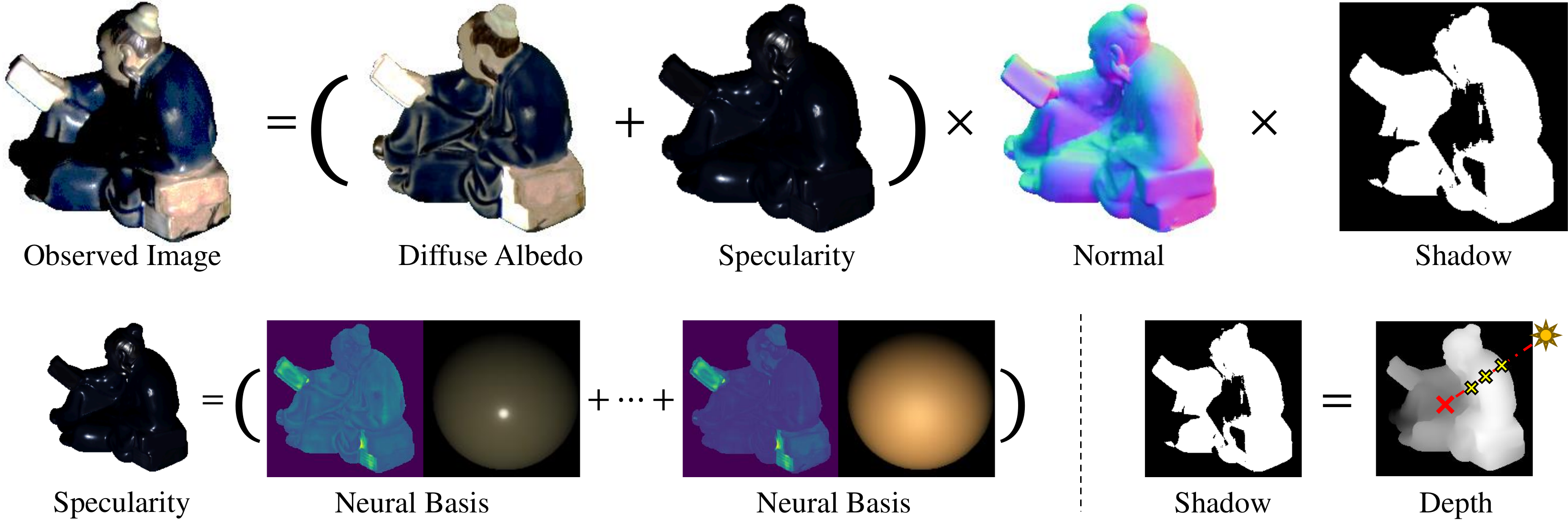}
\caption{\small We propose an self-supervised framework that estimates the surface normal, diffuse albedo, specularity, and shadow of an object. Our method learns the neural basis to fit the observed specularities accurately and gives clues for normal estimation. We also explicitly parameterize the shadows based on the estimated depth, alleviating artifacts on these shadows.}
\label{fig:overall_simple}
\end{figure*}



In this paper, we propose an unsupervised neural network method that overcomes the issues mentioned above. Our framework takes the image coordinates corresponding to a surface point as the input, and directly outputs the surface normal, reflectance parameters (\ie diffuse albedo and specular parameters), and depth at that surface point.  
We proposed a series of neural specular basis functions to account for the different types of specularities in the real-world. 
Our neural bases provide the parameterization for the surface reflectance and fit the object's appearance to obtain the accurate surface normal. Furthermore, our framework explicitly parameterizes the shadowed regions by tracing through the estimated depth map. These shadowed regions are then excluded from computation in order to avoid possible rendering artifacts.  Following the inverse graphics rendering idea, we use the estimated surface normal and neural reflectance to re-render the pixel intensities of the surface point under different light directions. Our framework is optimized by minimizing the difference between the reconstructed and observed images during the inference time. Therefore, there is no need for any ground truth data or pre-training. Our method outperforms both the supervised and self-supervised state-of-the-art methods on the challenging real-world dataset of DiLiGenT~\cite{shi2016benchmark}. Compared to other self-supervised deep methods~\cite{taniai2018neural, kaya2020uncalibrated}, our framework is ten times faster.

\section{Related Work}
\textbf{Conventional approaches:} 
The photometric stereo is firstly introduced by Woodham~\cite{woodham1980photometric}, which assumes the surface of the objects to be Lambertian and convex to avoid the specular effects and shadows. This problem can therefore be solved in a closed-form manner by least-squares. The above strict assumptions were gradually liberalized by later studies~\cite{wu2010robust,ikehata2012robust,mukaigawa2007analysis,wu2010photometric,queau2017non}. These methods can tolerate the existence of non-Lambertian effects by treating the specularities and cast shadows on the object as outliers.
However, 
they may also erase other clues specularities can bring.

\textbf{Supervised methods:} 
With the progress of deep learning in many of the computer vision areas, the learning-based methods are the ones that have achieved the best performance in photometric stereo recently~\cite{santo2017deep,ikehata2018cnn,li2019learning,chen2020deep,yao2020gps,wang2020non,zheng2019spline, honzatko2021leveraging, logothetis2021px}. 
Santo~\etal~\cite{santo2017deep} proposed the first network-based method, which per-pixelly estimates the normal by taking observed pixels in a pre-defined order. 
Chen~\etal~\cite{chen2018ps,chen2020deep} proposed a feature-extractor and features-pooling strategy to obtain the spatial information for photometric stereo. 
Recently, more works~\cite{yao2020gps,wang2020non} exploited the local and global photometric clues for this problem. 
These learning-based methods require a large amount of data with ground truth surface normal at the training stage. The synthesized data with some augmentation strategies are commonly used as collecting a large-scale real-world dataset is exceptionally expansive and impractical. 

\textbf{Self-supervised methods:} 
In contrast to the above-mentioned learning-based methods method, self-supervised methods do not require ground truth normal at supervision. Instead, the network is optimized by minimizing the difference between the reconstructed images and observed images. 
Taniai~\etal~\cite{taniai2018neural} proposed a self-supervised network that takes the whole set of images at the input, directly output the surface normal, and aiming to reconstruct the observed images. Their network structure is further expanded by Kaya~\etal~\cite{kaya2020uncalibrated} to deal with interreflection in the context of uncalibrated photometric stereo. Both of them implicitly encode specular components as features for the network and fail to consider shadows in the rendering equation. 

\begin{figure*}
\centering
\includegraphics[width=\textwidth]{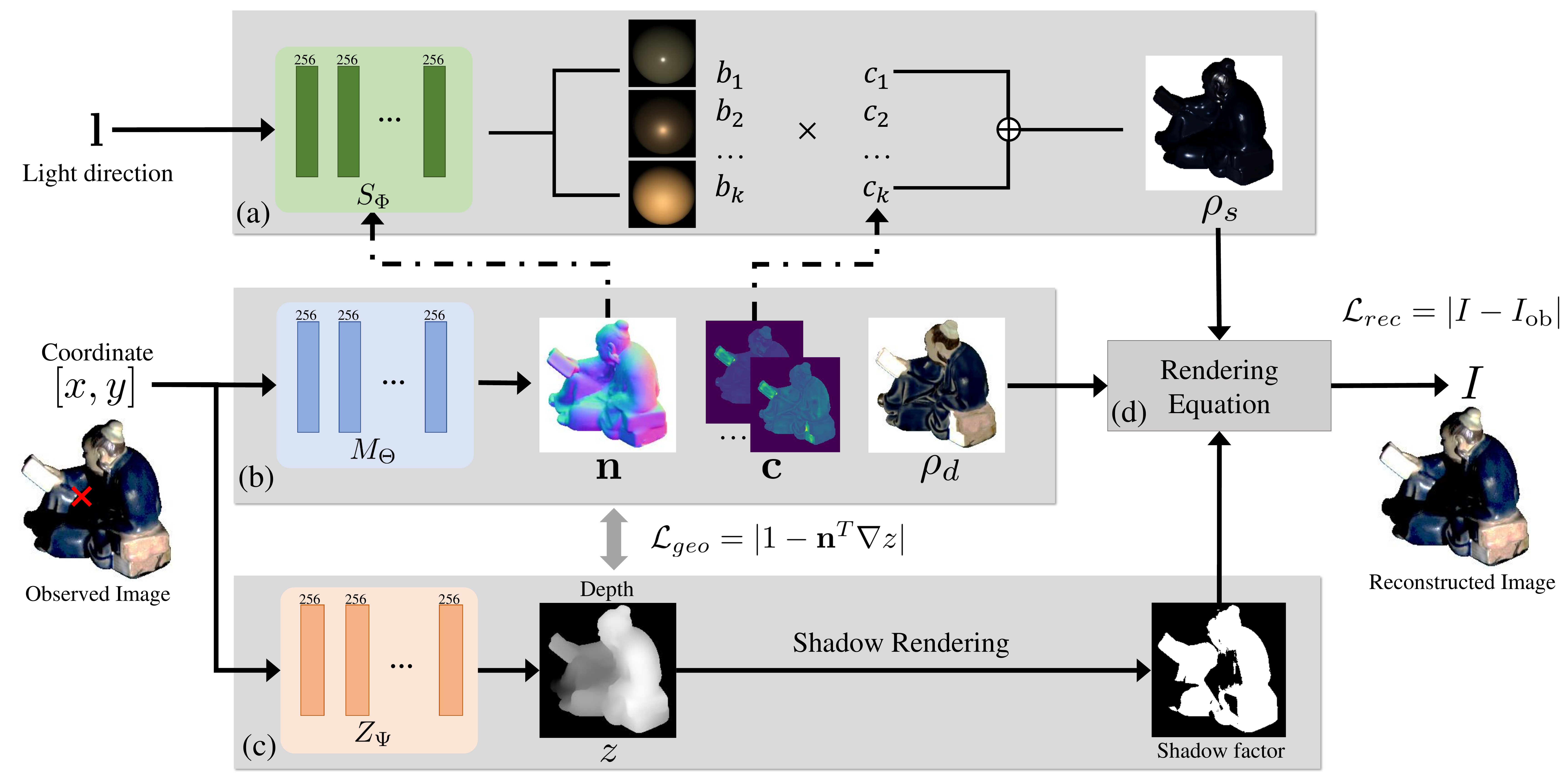}
\caption{\small The four modules of our MLP-based deep photometric stereo framework: (a) neural specular bases modeling $S_\Phi$ (see \sref{sec:brdf_modeling}) fits a suitable set of suitable BRDF bases to the target specularities; (b) surface modeling $M_\Theta$ (see \sref{sec:object_modeling}) estimates the surface normal, as well as parameters of the BRDF given the image coordinates as input; (c) $Z_\Psi$  estimates a dense depth map, which enables the shadow rendering (see \sref{sec:shadow_modeling}) by checking the visibility of the light source at each surface point; and (d) the rendering equation (see \sref{sec:rendering}). All  MLPs are optimized in a self-supervised manner by minimizing the reconstruction error between reconstructed and observed images.}
\label{fig:overall}
\end{figure*}

\textbf{Neural radiance fields:} 
Recently, neural radiance fields introduced by NeRF\cite{mildenhall2020nerf} is widely adopt in many reconstruction tasks in computer vision.
Many works also extend the neural radiance fields to  recover both the shapes and materials of the object~\cite{boss2020nerd, zhang2021nerfactor, zhang2021physg, srinivasan2021nerv}. 
These works are solving multi-view reconstruction problems. They generally assume the input being images of an object captured from multiple viewpoints under fix illumination.
In contrast, the photometric stereo problem we are focusing in this paper assumes multiple images taking from the same viewpoint, but with different illuminations.


\section{Proposed Method} \label{sec:method}
As shown in \fref{fig:overall_simple}, our framework aims at decoupling the surface into normal, diffuse albedo, specularity, and shadow. We model the specularity by learning a set of neural specular bases. Our method estimates the depth by querying the relative depth of the surface points. 
In the following subsections, we illustrate the details of each module in our framework. 

\subsection{Rendering Equation}\label{sec:rendering}

Following the conventional calibrated photometric stereo problem, we assume that the light source is in distance over the images with known light direction $\vl = [l_x,l_y,l_z]^T \in \mathcal{S}^2$ (the space of $3$-dimensional unit vectors)  and light intensity $L_i \in \mathbb{R}_+$. And the camera to be in orthographic position, hence, viewing direction $\vv=[0,0,-1]^T \in \mathcal{S}^2$. 
For simplicity, without any loss of generality, we omit the light intensity $L_i$ in the following formulations by dividing the observations (\ie images $I_i$)  with the corresponding lighting intensities, $I=I_i/L_i$. 
We also assume that there are no inter-reflections between the surfaces so that the point light source is the only light source to illuminate the target object. 

Given a light source from the direction $\vl$ illuminates a surface point with surface normal $\vn \in \mathcal{S}^2$. 
The observation $I$ viewing from direction $\vv $ can be written as 
\begin{align}\label{eq:image_model}
    I = s \rho(\vl, \vv, \vn) \max(\vl^T \vn, 0), 
\end{align}
where $s\in\{0,1\}$ is a binary variable with a value of $0$ at  shadows, and $1$ otherwise;
$\rho(\vl, \vv, \vn)$ represents the BRDF of the surface point, which is a function of the light, view direction, and the surface normal;  $\max(\vl^T \vn, 0)$ is the shading component. 


\subsection{Reflectance Modeling}\label{sec:brdf_modeling}
The Lambertian surface assumes the BRDF $\rho(\vl, \vv, \vn) = \rho_{\text{d}}$ is always a positive constant. This unrealistic assumption fails to account for those materials with high specular effects. It can be beneficial to model the specular part in BRDF and leverage its information for photometric stereo. In order to take both the diffuse and specular effects into account, here we choose a more realistic way to model the surface reflectance, \ie the microfacet BRDF models \cite{torrance1967theory, walter2007microfacet}, where the BRDF is separated into the diffuse and specular components
\begin{align}\label{eq:brdf}
    \rho(\vl, \vv, \vn) = \rho_{\text{d}} + \rho_{\text{s}}(\vl, \vv, \vn).
\end{align}


\begin{figure*}
\centering
\includegraphics[width=0.95\textwidth]{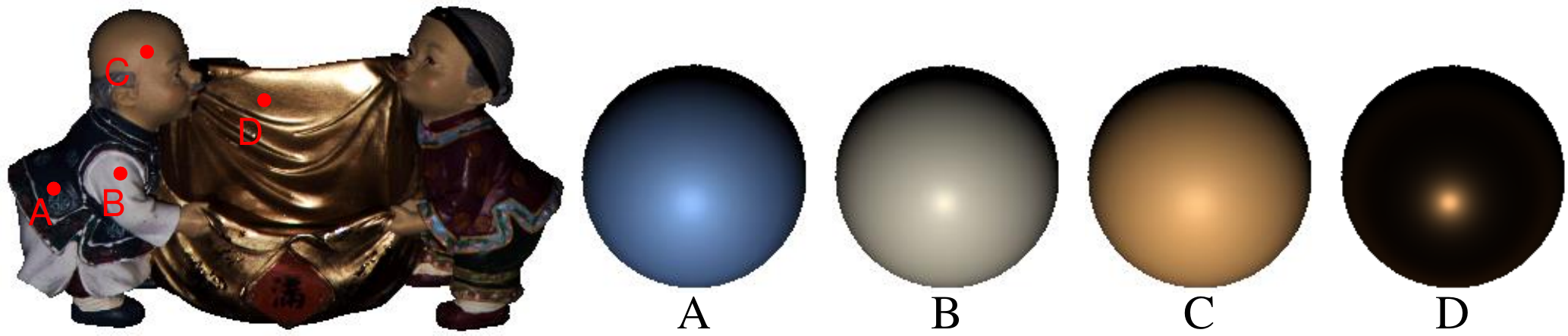}
\caption{\textbf{Visualization on the estimated svBRDFs.}  
We select four different surface points on the object ``Harvest'' and showcase our estimated BRDF spheres on the right.
The results demonstrate that our model can recover the metallic and diffuse materials. 
We scale up the observed images and normalize the BRDF spheres for better visualization. 
}
\label{fig:svbrdf_sphere_harvest}
\end{figure*}

\paragraph{Neural Specular Basis }
Previous deep-learning-based approaches implicitly handle the specularity on images by feeding them as features into their neural network~\cite{taniai2018neural,kaya2020uncalibrated}, or processed by max-pooling~\cite{chen2018ps,chen2020deep}. However, as the specularities, at the core, are  reflections on the surface, explicitly model these effects by using clues from physical reflection constraints will certainly bring merits to the photometric stereo problem.

To relieve the burden of fitting such a neural specular BRDF, we need to introduce some reasonable and realistic assumptions. Recalling that the BRDF can be converted to a half-vector $\vh$ based function with only four parameters~\cite{rusinkiewicz1998new}, we assume that our specular BRDF is isotropic and is only the function of half-vector $\vh$ and surface normal $\vn$. This assumption omits the Fresnel reflection coefficient and the geometric attenuation, which only has limited effects at grazing angles~\cite{burley2012physically}.
Besides, observing the fact that many surface points in the real-world object are similar, if not identical, in the material. We further assume that the specular BRDF $\rho_{\text{s}}(\vl, \vv, \vn)$ at each surface point lies on a non-negative linear combination of the atoms of specular basis. Similar approaches for simplifying the BRDF model to be the combination of different bases were also used in previous works~\cite{Matusik:2003,hui2017shape}.
The specular BRDF can then be written as
\begin{align}
\rho_{\text{s}}(\vl, \vv, \vn) = \mathbf{c}^T \ D(\vh, \vn)  , \quad \vh=\dfrac{\vl + \vv}{||\vl + \vv||} ,
\end{align}
where $\vh$ is the half-vector between lighting and viewing direction; and $D(\vh, \vn) = [b_1, b_2, \cdots, b_k]^T$ is the underlying specular basis of the target object; $\left[ c_1, c_2,\cdots,c_k \right]^T  := \mathbf{c} \in \mathbb{R}_+^k$ represent the weights of each specular basis; $k$ is the number of different bases. 
We assume that $\mathbf{c}$ is an element-wise non-negative  vector, suggesting that the surface reflectance is represented by positive combination of a small number of basis materials. 

We use an MLP to parameterize the specular basis by
\begin{align}
D(\vh, \vn) = S_\Phi(\vh, \vn) ,
\end{align}
The network $S_\Phi(\vh, \vn)$ only takes  $\vh, \vn$ at the input,  outputs the different specular basis in form of $[b_1, b_2, \cdots, b_k]^T$, as shown in \fref{fig:overall}. $\Phi$ are its weights that can be optimized during testing.
It is well established that a variety of reflectance maps can be represented by a linear combination of a few basis functions~\cite{hertzmann2005example,matusik2003efficient,Matusik:2003}.
We empirically set $k=9$ when testing our model on real datasets. 
In \fref{fig:svbrdf_sphere_harvest}, we re-rendered several spheres by using our estimation on the reflectance and neural basis of the surface points.
As shown in \fref{fig:svbrdf_sphere_harvest}, our neural reflectance modeling can approximate the spatially-varying and non-Lambertian materials very well. It can recover the diffuse surface, and also reliably construct the high-peak and long-tail metallic specularites.

\subsection{Surface modeling}\label{sec:object_modeling}
We model the surface normal, diffuse, and neural basis coefficients of an object by an MLP  $M_\Theta$. It takes the image coordinates of the pixels $\vx = [x, y]^T \in \mathbb{R}^2$ as input. The output is the corresponding surface normal $\vn$, diffuse albedo $\rho_{\text{d}}$, and the coefficients $\mathbf{c}$ of the bases at each coordinate $\vx$.
\begin{align}\label{eq:normal_net}
   \vn, \rho_{\text{d}}, \mathbf{c} =  M_\Theta(\vx) ,
\end{align}
where $\mathbf{c}$ represents the coefficients that can be used to reconstruct the specular component $\rho_{\text{s}}$ in \sref{sec:brdf_modeling}; and $\Theta$ is the weights of this MLP that can be optimized.

We use a similar MLP architecture and positional encoding strategy from NeRF~\cite{mildenhall2020nerf} to build our network, and the embedding in input coordinates $\vx$. The difference is that while NeRF also takes different viewing directions as input to model the view-dependent effects of the objects' appearance, our $M_\Theta$ network only estimates the ``static'' properties of the target object. Instead, we cover the ``light-dependent'' variance of the object by neural reflectance modeling. Our design will encourage the network to correctly decompose surface normal and material property of the object.

\begin{figure*}
\centering
\includegraphics[width=0.9\textwidth]{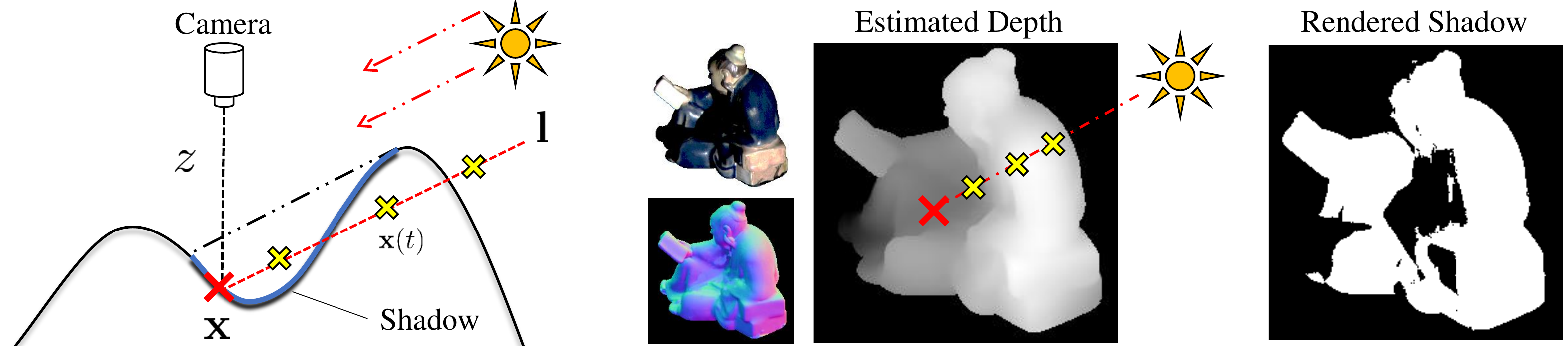}
\caption{\small Shadow parameterization and rendering. As shown in the left figure, shadows are caused by self-occlusion. To determine whether a surface point $\vx$ falls into the shadow region, we trace the point to the light source and sample multiple points $\vx(t)$ along this ray. Given the light direction $\vl$ and the estimated depth map $Z_\Psi(\vx)$, we can query the depth and compare the values to effectively parameterize and render the shadow by \eref{eq:shadow_rendering}. }
\label{fig:shadow_render}
\end{figure*}

\subsection{Shadow handling}\label{sec:shadow_modeling}

We now look at the shadow factor $s$ in the image rendering \eref{eq:image_model}. 
Due to the rugged surface of the objects in the world, shadows may appear at the reflecting surface. As shown in \fref{fig:shadow_render}, shadow occurs when the object itself occludes the surface. 
Rendering of the shadowed region relies on the relative geometry and depth of the object with respect to the light directions. 
Hence, we introduce a depth MLP $Z_\Psi$ to model the object's depth value $z\in \mathbb{R}$ between the object surface points to the camera. The depth MLP takes images coordinates as input, outputs the corresponding depth value of the given coordinates $z = Z_\Psi(\vx).$

To examine whether the object occludes the light source and hence causing the shadow, we can draw a line from the surface point $\vx$ toward the light source. Denote this line in the world coordinates as 
$\mathbf{L} = \mathbf{X} - t \vl , $ where $  t\in (0,+\infty)$; the $\mathbf{X} = [x, y, z]$ represent the surface points with its depth value $z$ given by $Z_\Psi(\vx)$. We can further simplify the equation by using the function $L_z$ to denote the $z$-axis value of $\mathbf{L}$. Now, by traveling along the light direction, \ie $t\in(0,+\infty)$, we can compute the shadow factor by 
\begin{align}\label{eq:shadow_rendering}
    s = \step \big( \min_{\vx(t)}  \left( Z_\Psi(\vx(t)) - L_z(\vx(t)) \right) \big), \quad  \vx(t) = \vx - t \vl' ,
\end{align}
where the $\step(\cdot)$ denote the Heaviside step function, which outputs $1$ if input is positive, and $0$ otherwise; $\vl'=[l_x,l_y]^T$ is the projection of light direction $\vl$ at $xy$-plane. 
In implementation, we set the step size for shadow rendering to be 32 (with logspace intervals). 


\section{Implementation}
We use the positional encoding~\cite{mildenhall2020nerf} strategy to encode the input before inputting them into the MLP.  For surface modeling net $M_{\Theta}$, we encode the input with 10 levels of Fourier functions, the network $M_{\Theta}$ uses 12 fully-connected ReLU layers with 256 channels. The surface normal $\vn$ is output at $8$-th layer while the BRDF parameters are output at the last layer. 
We also use 10 encoding functions to embed the input of depth net $Z_\Psi$, which has 8 fully-connected ReLU layers with 256 channels. 
For the neural basis MLP $S_\Phi$, we use only 3 encoding functions to embed the input. The network $S_\Phi$ consists of 3 fully-connected ReLU layers with 64 channels.  Please refer to the supplementary material for more implementation details. 
Overall, the three MLP-networks are rather lightweight (\ie small footprint) with total combined parameters of merely $1.1$M.  In contrast, the CNN-based self-supervised method~\cite{kaya2020uncalibrated} contains $3.7$M parameters. Besides, our model is shallow and require less computation than previous works. Hence, our framework is much faster in inference time.
The inference time in the 10 objects of DiLiGenT dataset range from 3 min to 9 min, with an average of 6 min per object. In contrast, CNN-based methods~\cite{taniai2018neural,kaya2020uncalibrated} took about an hour  per object.

\textbf{Reconstruction loss.}
The reconstruction loss is defined as mean absolute errors between the observed intensity $I_{\text{ob}}$ and reconstructed intensity:
\begin{align}
    \mathcal{L}_{rec} = \sum_{\text{all pixels}} |I - I_{\text{ob}}| .
\end{align}

\begin{table*}
\centering
\caption{Quantitative comparison on the DiLiGenT dataset. The metric here is mean angular error (MAE); the lower MAE is preferred.}
{\small
\begin{tabular}{@{}c|c|ccccccccccc@{}}
\toprule
GT normal & Methods & Ball          & Bear          & Buddha        & Cat           & Cow           & Goblet        & Harvest        & Pot1          & Pot2          & Reading       & Avg.          \\ \midrule \hline
No        & Ours    & 2.43          & \textbf{3.64} & \textbf{8.04}          & \textbf{4.86}          & \textbf{4.72} & \textbf{6.68} & \textbf{14.90}          & \textbf{5.99}          & \textbf{4.97} & \textbf{8.75} & \textbf{6.50} \\
No        & TM18\cite{taniai2018neural}    & \textbf{1.47} & 5.79          & 10.36         & 5.44          & 6.32          & 11.47         & 22.59          & 6.09          & 7.76          & 11.03         & 8.83          \\ 
No        & BK21\cite{kaya2020uncalibrated}                                      & 3.78          & 5.96          & 13.14         & 7.91          & 10.85         & 11.94         & 25.49          & 8.75          & 10.17         & 18.22          & 11.62         \\
No        & L2\cite{woodham1980photometric}      & 4.10          & 8.40          & 14.90         & 8.40          & 25.60         & 18.50         & 30.60          & 8.90          & 14.70         & 19.80         & 15.40         \\\hline
Yes       & PX-NET\cite{logothetis2021px} & 2.00 & \textbf{3.50} & 7.60 & \textbf{4.30} & \textbf{4.70} & \textbf{6.70} & \textbf{13.30} & \textbf{4.90} & \textbf{5.00} & \textbf{9.80} & \textbf{6.17} \\
Yes       & WJ20\cite{wang2020non}    & \textbf{1.78}          & 4.12          & \textbf{6.09} & 4.66          & 6.33          & 7.22          & 13.34 & 6.46          & 6.45          & 10.05         & 6.65          \\
Yes       & CNN-PS\cite{ikehata2018cnn}  & 2.20          & 4.10          & 7.90          & 4.60 & 8.00          & 7.30          & 14.00          & 5.40 & 6.00          & 12.60         & 7.20          \\
Yes       & GPS-Net\cite{yao2020gps} & 2.92          & 5.07          & 7.77          & 5.42          & 6.14          & 9.00          & 15.14          & 6.04          & 7.01          & 13.58         & 7.81          \\
Yes       &PS-FCN\cite{chen2018ps}	& 2.82	& 7.55	& 7.91	& 6.16	& 7.33	& 8.60	& 15.85	& 7.13	& 7.25	& 13.33	& 8.39\\ \bottomrule
\end{tabular}
}
\label{tab:all}
\end{table*}

\begin{figure}
\centering
\includegraphics[width=0.5\textwidth]{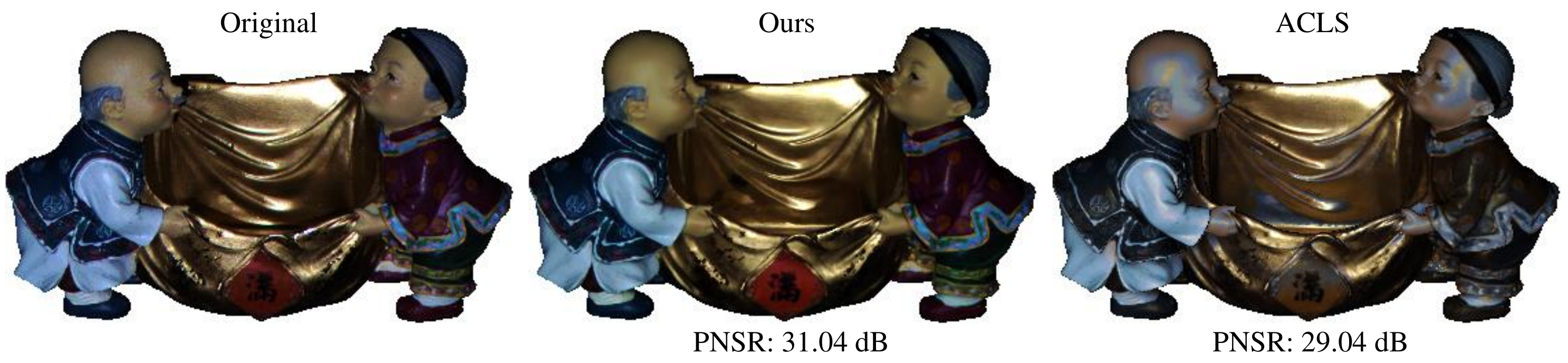}
\caption{\small Re-rendered image by our estimated svBRDFs. From left to right, we showcase the original image captured from ``Harvest'', the re-rendered image using our estimated neural svBRDFs, and the re-rendered image by ACLS~\cite{alldrin2008photometric}, respectively. 
Our method achieves a better quality in reconstruction, being $2$dB higher in peak signal-to-noise ratio (PSNR).
ACLS failed to recover the spatially-varying materials (the red cloth and the human faces are all fainted in ACLS's result).
}
\label{fig:recon_img}
\end{figure}


\textbf{Geometry Constraint.}
We introduce a geometry constraint between the estimated surface normal $\vn$ and depth network $Z_{\Psi}$ as below
\begin{align}
    \mathcal{L}_{geo} = \sum_{\text{all pixels}} (1 - \vn^T \nabla Z_{\Psi}) .
\end{align}
In the early stage of optimizing the network $Z_{\Psi}$, 
we introduce shadow guidance $s_g$ to help with the training.
Assume that observation under $n$ different light direction is $[I_{1},I_2,...,I_n]$. We then set a threshold as $ 0.1\lambda_m$, where $\lambda_m = \frac{1}{n} \sum I_{i}$ is the mean intensity. 
Those pixel intensities that are smaller than the threshold will be discard.
We use \eref{eq:shadow_rendering} for shadow rendering once the depth network $Z_{\Psi}$ is stable.

\textbf{Smoothness constraint.}
Previous self-supervised methods suffered from poor network initialization~\cite{taniai2018neural,kaya2020uncalibrated}. Their networks required a pre-computed surface normal map as the early network guidance. In contrast, our model does not need any pre-computed surface normal as guidance. Instead, to cope with the poor network initialization problem, we use a smoothness constraint to guide the network in the early stages since the albedo and surface normal of real-world objects usually present a piece-wise smooth pattern
\begin{align}
    \mathcal{L}_{tv} = V_{l_1}(\rho_{\text{d}}, \mathbf{c}) + V_{l_2}(\vn) ,
\end{align}
where $V_{l_1}$ represents the total variation function with absolute loss and $V_{l_2}$ with square loss.

To sum up, we optimize the parameters of the MLPs $M_\Theta, S_\Phi, Z_{\Psi}$ by minimizing the following loss function: $\mathcal{L} = \mathcal{L}_{rec} + \mathcal{L}_{geo} + \beta \mathcal{L}_{tv}$, where $\beta$ is the hyper-parameter controlling the total variation loss. We set it as $\beta=0.01$; and it will then be set to $0$ after the first half iterations.

\section{Experiments} \label{sec:experiments}
In this section, we evaluate our method and its variants on the challenging real-world dataset DiLiGenT~\cite{shi2016benchmark}. We used all the $n=96$ images under different light directions for optimizing the network, except the object ``Bear''. We discard the first $20$ images of ``Bear'', as they are found to be over-saturated in previous work~\cite{ikehata2018cnn}. The batch size is set as $8$ images per batch. We iterate in total $6000$ iterations when optimizing the network. We use Adam~\cite{kingma2014adam} optimizer with a learning rate of $5\times10^{-4}$ and other parameters at their default settings.
Our method is implemented in PyTorch and is running on a RTX 3090 GPU.
The inference (\ie training) time in the 10 objects of DiLiGenT dataset range from 3 min to 9 min, with an average of 6 min per object. In contrast, previous CNN-based methods \cite{taniai2018neural,kaya2020uncalibrated} took about an hour  per object.

We also evaluate our method in two other challenging real world datasets: Gourd\&Apple dataset~\cite{alldrin2008photometric}, and Light Stage Data Gallery~\cite{chabert2006relighting}. Please refer to supplementary material for more details.

\begin{figure*}
\centering
\includegraphics[width=0.95\textwidth]{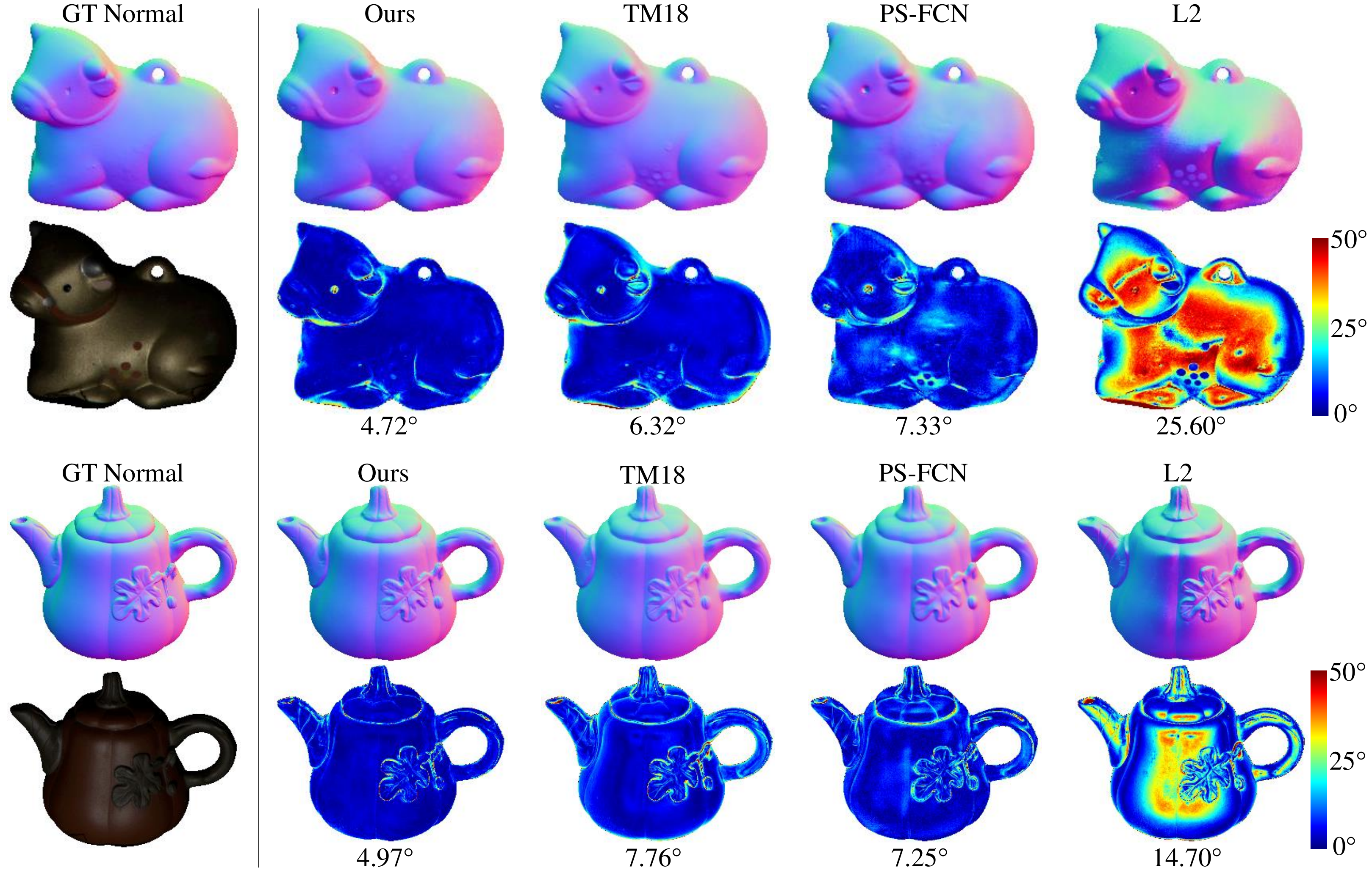}
\caption{\small Qualitative results on ``Cow'' and ``Pot2''. For each object, the odd numbered rows show the observed image and estimated normal by different methods; the even numbered  rows show the angular (normal) error in degrees by different methods.}
\label{fig:all}
\end{figure*}

\subsection{Evaluation on real-world dataset}\label{sec:exp_real}

\textbf{Surface normal evaluation.} In \Tref{tab:all}, we present the quantitative comparison of our method against other methods on the DiLiGenT dataset. We use the mean angular error (MAE) as the metric in the paper. The lower MAE is preferred. We classify the previous methods into two categories: the supervised methods, which need ground truth surface normal at the training stage; and the self-supervised which does not need ground truth surface normal and directly estimates the normal at testing time. 
As reported in the \Tref{tab:all}, our method achieves the best performance over the other self-supervised methods at average MAE errors. Comparing to the previous self-supervised method~\cite{taniai2018neural,kaya2020uncalibrated}, our method is $2.33$ degrees better in MAE errors.
Thanks to our neural reflectance modeling, our method shows its significant advantages on shiny objects like ``Reading'', ``Cow'' and ``Goblet''. 
We present the visualization of ``Cow'' and ``Pot2'' in \fref{fig:all}. ``Cow'' is a typical metallic-painted object with a high peak of specularities; while ``Pot2'' shows more broad and soft specular effects. Our method achieves the best performance in both two cases.

\textbf{svBRDF evaluation.} In \fref{fig:svbrdf_sphere_harvest}, we visualize the estimated svBRDFs on the challenging object ``Harvest''. ``Harvest'' contains many different type of materials over the surface. From diffuse (see point A), to specular (see point D), our model presents visually pleasing estimated BRDF spheres over these different points. 
To quantitatively evaluate our method, we re-rendered the observed image with our estimated reflectances, and ground truth lights. The results are shown in \fref{fig:recon_img}. 
We compare our re-rendered images with ACLS~\cite{alldrin2008photometric}.
ACLS's BRDF fitting results are provided by Shi~\etal~\cite{shi2016benchmark}, where it takes the ground truth surface normal when fitting the BRDF. 
By looking at the re-rendered images, our method achieve much higher reconstruction quality ($2$dB higher in peak signal-to-noise ratio (PSNR)). In comparison, ACLS~\cite{alldrin2008photometric} failed to faithfully recover the spatially-varying materials.


\begin{figure*}
\centering
\includegraphics[width=0.95\textwidth]{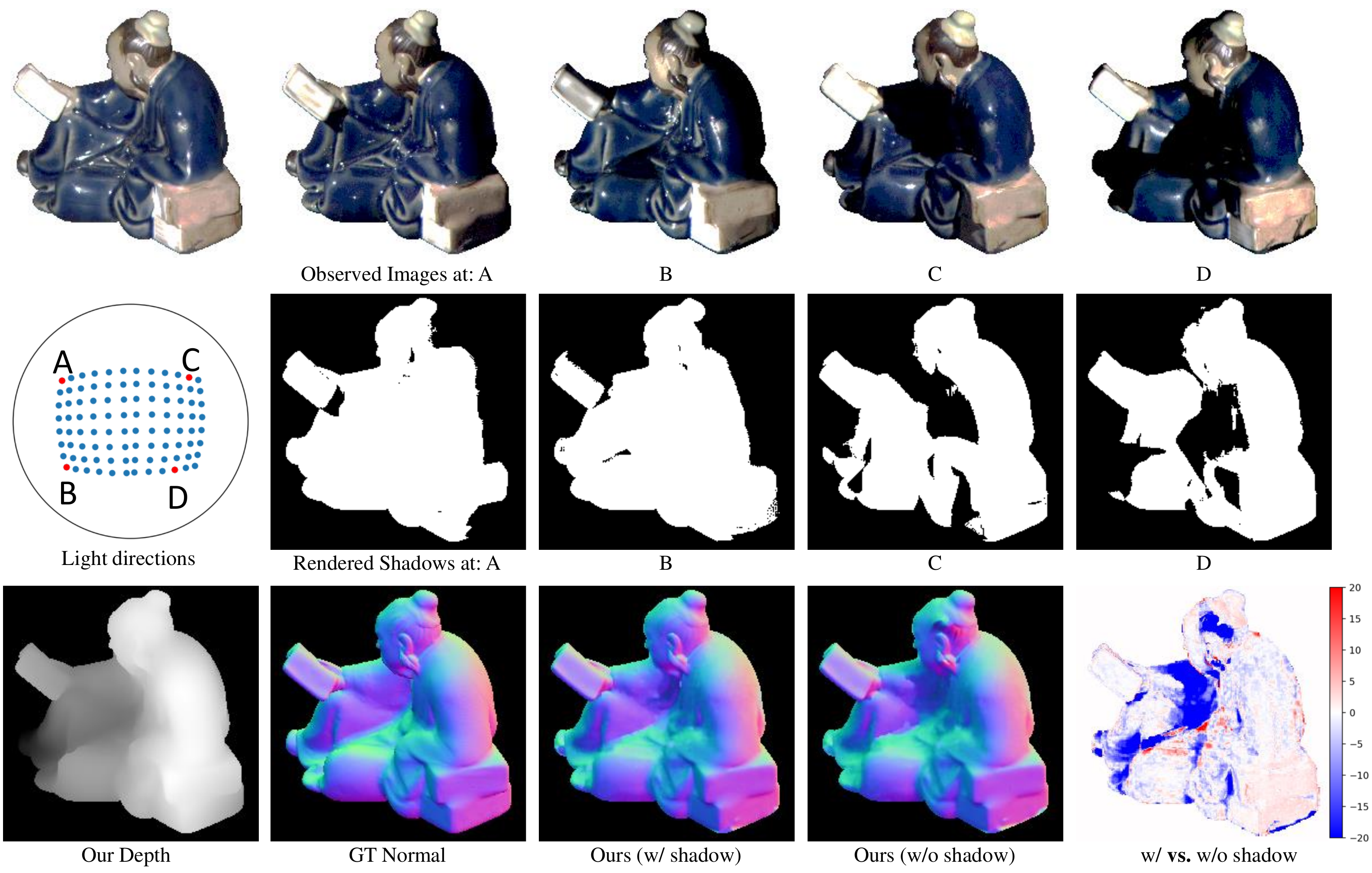}
\caption{\small We select 4 different light directions. Their distribution is labeled as red points in the light distributions image in the second row. 
The first row shows the observed images under these 4 different light sources. The second row presents the results of our rendered shadow region under the corresponding illuminations. In the third row, we showcase the estimated depth, ground truth surface normal, estimated surface normal (with and without the shadow factor). 
In the right-most image on the third row, we also compare our estimated normal ``w/ shadow'' and ``w/o shadow''. The blue color in the comparison corresponds to the area where ``w/ shadow'' outperforms ``w/o shadow''.
}
\label{fig:shadow_exp}
\end{figure*}

\begin{table}
\centering
\caption{Quantitative results on DiLiGenT with different number of images at the input. The average MAEs are shown in table.}
\scriptsize
\begin{tabular}{@{}c|c|cccc@{}}
\toprule
GT Normal & \# inputs           & 96   & 16   & 10    & 8     \\ \midrule
No        & Ours                & 6.50 & \textbf{6.82} & \textbf{7.47}  & \textbf{7.70}  \\
Yes       & LMPS~\cite{li2019learning}       & 8.43 & 9.66 & 10.02 & 10.39 \\
Yes       & PX-Net~\cite{logothetis2021px}     & \textbf{6.17} & --   & 8.37  & --    \\
Yes       & SPLINE-NET~\cite{zheng2019spline} & --   & --   & 10.35 & --    \\ \bottomrule
\end{tabular}
\label{tab:smallinputs}
\end{table}

\textbf{Results with Sparse Inputs.}
To evaluate how the performance changes with a different number of images at the input, we test our method on the DiLiGenT dataset. We follow the previous work LMPS~\cite{li2019learning} to use the same inputs for our method. The results and comparison are presented in \cref{tab:smallinputs}. From left to right, our method takes 96 images, 16 images, 10 images and 8 images at the input separately. 
To our best knowledge, the trained model of SPLINE-Net~\cite{zheng2019spline} and PX-Net~\cite{logothetis2021px} is not publicly available. Hence, we report the value from their original paper. 
Although our method is not designed for sparse inputs, we still perform significantly better than previous work under a small number of inputs. It demonstrates that our method is  robust to the sparse inputs.  


\subsection{Ablation Study}\label{sec:exp_shadow}
\textbf{Shadow handling: }To show the efficacy of our shadow handling mechanism, we conduct ablation study by removing the shadow rendering module, denote as ``w/o shadow''.  Quantitative comparisons are shown in \Tref{tab:shadow}, where one can see that the mean angular error on all objects increases $1.96$ degrees. Notably, the performance degradation is majorly caused by objects ``Buddha'', ``Harvest'' and ``Reading''. This is as expected, because these objects have rather complex (concave) surface geometry, more susceptible to cast shadows. Our proposed shadow handling method attends to these shadowed regions better, achieving high recovery accuracy. In \fref{fig:shadow_exp}, we give the visualization of the effects of our shadowing module on the object ``Reading''. Observing the image and its ground truth normal of this object, we can see that ``Reading'' is a highly non-convex object with many specularities and shadows. The shadowed region is especially big when the light comes from the right direction, as shown in the lighting direction C and D in the figure. Our render shadows under these lighting directions, despite some minor errors, accurately predicting the shadowed regions.  The error map shown at the right-most of the third row in \fref{fig:shadow_exp} corresponds to the difference between the MAE yielded by our proposed model and its no-shadow-variant (``w/o  $s$''). The negative areas, \ie, blue regions in the error map, are those where our proposed model outperforms the alternative. The full model performs better in the region where the shadows are evident.

\begin{table}
\centering
\caption{\small Evaluations of the different variants of the proposed method. The second row is  without using the early stage smoothness constraint; the third row is the method without the shadow factor $s$; the last row is without using specular component $\rho_s$. The metric here is MAE; lower is preferred.}
{\scriptsize
\setlength{\tabcolsep}{1.4pt}
\begin{tabular}{@{}l|ccccccccccc@{}}
\toprule
Methods    & Ball & Bear & Buddha & Cat  & Cow  & Goblet & Harvest & Pot1 & Pot2 & Reading & Avg. \\ \midrule
Proposed & 2.43 & \textbf{3.64} & \textbf{8.04}   & \textbf{4.86} & \textbf{4.72} & \textbf{6.68}   & \textbf{14.90}   & \textbf{5.99} & \textbf{4.97} & \textbf{8.75}   & \textbf{6.50} \\
w/o $\mathcal{L}_{tv}$     & 2.44 & 3.66 & 8.56   & 4.93 & 5.27 & 6.77   & 21.67   & 6.73 & 6.88 & 9.19    & 7.61 \\ 
w/o $s$ & \textbf{2.13} & 4.29 & 11.09  & 6.81 & 5.69 & 8.30   & 17.88   & 7.79 & 7.80 & 12.68   & 8.44 \\
w/o $\rho_s$ & 3.13 & 6.48 & 10.58 & 6.93 & 27.23 & 15.19 & 29.65 & 8.27 & 14.14 & 11.41 & 13.30 \\\bottomrule
\end{tabular}
}
\label{tab:shadow}
\end{table}

\textbf{Effectiveness of smoothness constraint:}
To show the effectiveness of proposed smoothness constraint, we conduct the experiments without using this loss, denote as ``w/o $\mathcal{L}_{tv}$'', shown in \Tref{tab:shadow}. The mean angular error on average is $1.11$ degrees lower by leveraging this constraint. 

\textbf{Effectiveness of specular modeling:}
We further test the model without using any specular modeling, denote as ``w/o $\rho_s$'', shown in \Tref{tab:shadow}.
The performance is significantly worse without using the specular $\rho_s$. We can see that with specular components, our method improves a lot for the shiny objects like ``Cow'', ``Goblet'' and ``Harvest''.

\section{Discussions and Conclusions}\label{sec:discussions}
In this paper, we have proposed an MLP-based approach for non-Lambertian shape reconstruction. The key novelty of our method is the neural parameterizations of spatially-varying surface reflectances, and of surface geometry. By leveraging the physical principle of image rendering, we explicitly tackle the reflectance and cast shadows by neural network.  Despite being an unsupervised method, our method outperforms existing state-of-the-art supervised methods on real-world datasets. Our method is inspired by NeRF~\cite{mildenhall2020nerf}, which uses a coordinate-based MLP to model the mapping from 3D coordinates to appearance. In contrast, we factorize the image appearance into multiple components: normal, diffuse albedos, neural specular bases, and shadows. The fitting on these physical-based rendering factors restores the object's surface properties faithfully. Besides, we explicitly parameterize diffuse, specularities, and shadows to ensure the inverse rendering follows a physically meaningful and explainable manner. 
Our method also relates to \cite{taniai2018neural,kaya2020uncalibrated}, which aim at optimizing a CNN-based self-supervised architectures. Our MLP-based framework is significantly faster than those CNN-based methods. We will release the code and models. 

\textbf{Limitations and future work:} 
Our estimation of depth is sensitive to the accuracy of normal estimation and surface discontinuities. Introducing more constraints for accurate depth estimation would certainly help to identify more accurate shadows. 
Our model may fail in the presence of strong inter-reflections. Finding an efficient model to trace secondary and tertiary rays bouncing between surfaces is also an interesting future direction.


\textbf{Acknowledgments}
This research is funded in part by ARC-Discovery grants (DP 190102261 and DP220100800), a gift from Baidu RAL, as well as a Ford Alliance grant to Hongdong Li.

{\small
\bibliographystyle{ieee_fullname}
\bibliography{iclr2022_conference}
}

\newpage
\onecolumn
\appendix
\newpage
\section*{Supplementary material}

\section{Implementation Details}\label{sec:appendix_implementation}

\begin{figure*}[!h]
\centering
\includegraphics[width=0.7\textwidth]{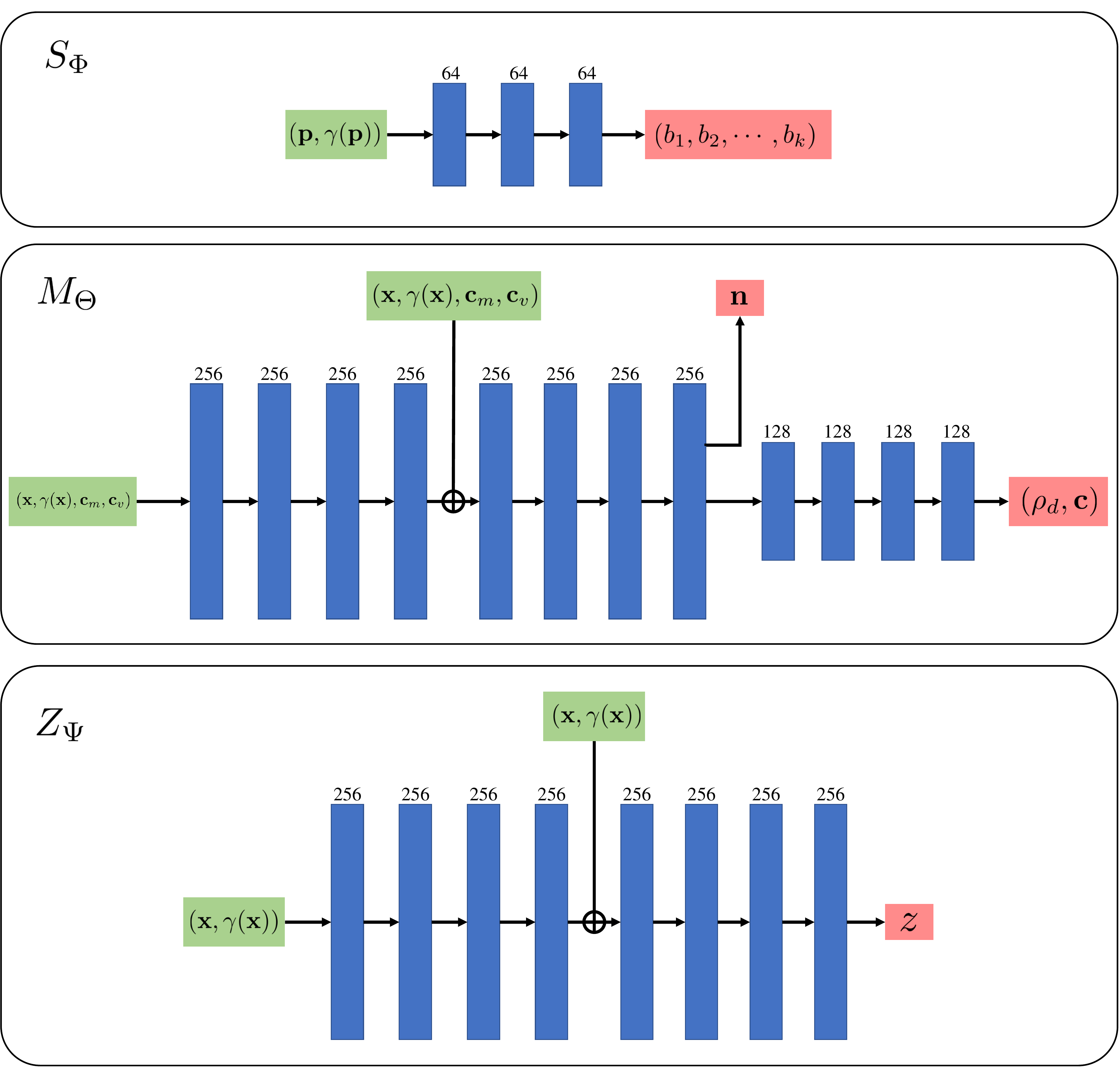}
\caption{\textbf{Network architecture of our three MLPs:} $S_\Phi, M_\Theta, Z_\Psi$. In this figure, inputs of the network are shown in the green blocks; outputs are shown in the red blocks. The blue blocks represent the fully-connected layers with its size of the hidden channels stated on the top. All fully-connected layers are followed by a ReLU activation layer, except the output layers. The ``$\oplus$'' in the middle of the $M_\Theta, Z_\Psi$ network denotes the vector concatenation: we add a skip connection after the fourth layer of $M_\Theta, Z_\Psi$, and concatenate its output features with the input. }
\label{fig:network}
\end{figure*}

\paragraph{Network architecture}
In \fref{fig:network}, we show the detail structure of our three MLPs : (a) specularity basis modeling $S_\Phi$; (b) surface modeling $M_\Theta$, (c) depth modeling $Z_\Psi$. The design of these structures are inspired by a recent work NeRF~\cite{mildenhall2020nerf}.

\paragraph{Positional encoding of input of $M_\Theta$ and $Z_\Psi$}
The inputs of surface modeling $M_\Theta$ and depth modeling $Z_\Psi$ are the pixel coordinates $\vx$. 
We adopt the positional encoding strategy~\cite{mildenhall2020nerf} to embed the input coordinates $\vx\in \mathbb{R}^2$ into a higher space $\vx\in \mathbb{R}^{4m}$:
\begin{align}
    \gamma(\xi) = (&\sin (2^0 \pi \xi), \cos(2^0\pi \xi), \cdots, \sin (2^{m-1} \pi \xi), \cos(2^{m-1}\pi \xi)).
\end{align}
In practice, we first normalized the coordinates to range $(-1,1)$, then apply the above encoding function with $m=10$ to each of the two coordinate values in $\vx$. Then, we concatenate  the coordinate and its embeddings as $(\vx, \gamma(\vx))$ to be  the input of the two MLPs.

\paragraph{Color mean and variation for $M_\Theta$ albedo estimation}
Since the albedo estimation of an object surface exists an ambiguities in its scales. We compute the mean value and variation value $\mathbf{c}_m,\mathbf{c}_v$ of the images and concatenate them to the input of $M_\Theta$ for albedo estimation.

\paragraph{Positional encoding of input of $S_\Phi$}
Rusinkiewicz~\cite{rusinkiewicz1998new} reparameterized the BRDF as a function of the half-vector $\vh$ (\ie  the half-vector between lighting and viewing direction). This half-vector-based parameterization is further evaluated and discussed
by \cite{pacanowski2012rational,burley2012physically}, 
where they found that a simplified isotropic BRDF can be modeled in two parameters $(\theta_h, \theta_d)$, where $\theta_h = \arccos(\vn^T \vh) ,  \theta_d = \arccos(\vv^T \vh)$.
In our method, we take the cosine value of these two variables as the input:
\begin{align}
    \mathbf{p} = (\vn^T \vh , \vv ^T \vh).
\end{align}
Then, the input $\mathbf{p}$ is further encoded by $\gamma (\mathbf{p})$ with $m=3$. Likewise, we concatenate $\mathbf{p}$ and its embeddings as $(\mathbf{p}, \gamma(\mathbf{p}))$ to be the input of the specularity basis modeling $S_\Phi$.



\section{Additional ablation study }
In this section, we compare a variant of our surface modeling network $M_\Theta$.
The original model we use in the paper is denoted as the baseline. The equation for baseline surface modeling (i.e. the one used in main paper) is: $\vn, \rho_{\text{d}}, \mathbf{c} =  M_\Theta(\vx)$.
\textbf{Surface Modeling Variant 1}: directly output depth, diffuse albedo and specular weights rather than outputting normals.
\begin{align}
   z, \rho_{\text{d}}, \mathbf{c} =  M^1_\Theta(\vx) ,
\end{align}
\begin{table}[h]
\centering
\caption{Quantitative comparison on different variants of surface modeling. The metric here is MAE; lower is preferred. Below, we present the average MAE of ten objects in DiLiGenT.}
\small
\begin{tabular}{@{}c|c @{\hskip 5mm}}
\toprule
\qquad Methods \qquad       & \qquad Avg.    \qquad      \\ \midrule
\qquad Baseline \qquad    & \qquad  \textbf{6.50} \qquad \\
\qquad Variant 1 \qquad  & \qquad 7.56 \qquad \\
\bottomrule
\end{tabular}
\label{tab:variant}
\end{table}

Our experiments in \Tref{tab:variant} show that the baseline model (outputting normals, albedo and specular weights by a single network) has the best performance.
Our analysis is that:
Comparing to \textbf{Variant 1}: The surface normal is closely related to the photometric appearance of an object. By directly outputting normal, our baseline network can achieve a lower photometric appearance (reconstruction) loss. If we directly output depth, we need to apply an additional step to compute the finite difference to get normal. Hence, directly outputting normal will help the training of the network to minimize the reconstruction loss.


\newpage
\section{Evaluation on re-rendered images}
\begin{figure*}[!h]
\centering
\includegraphics[width=\textwidth]{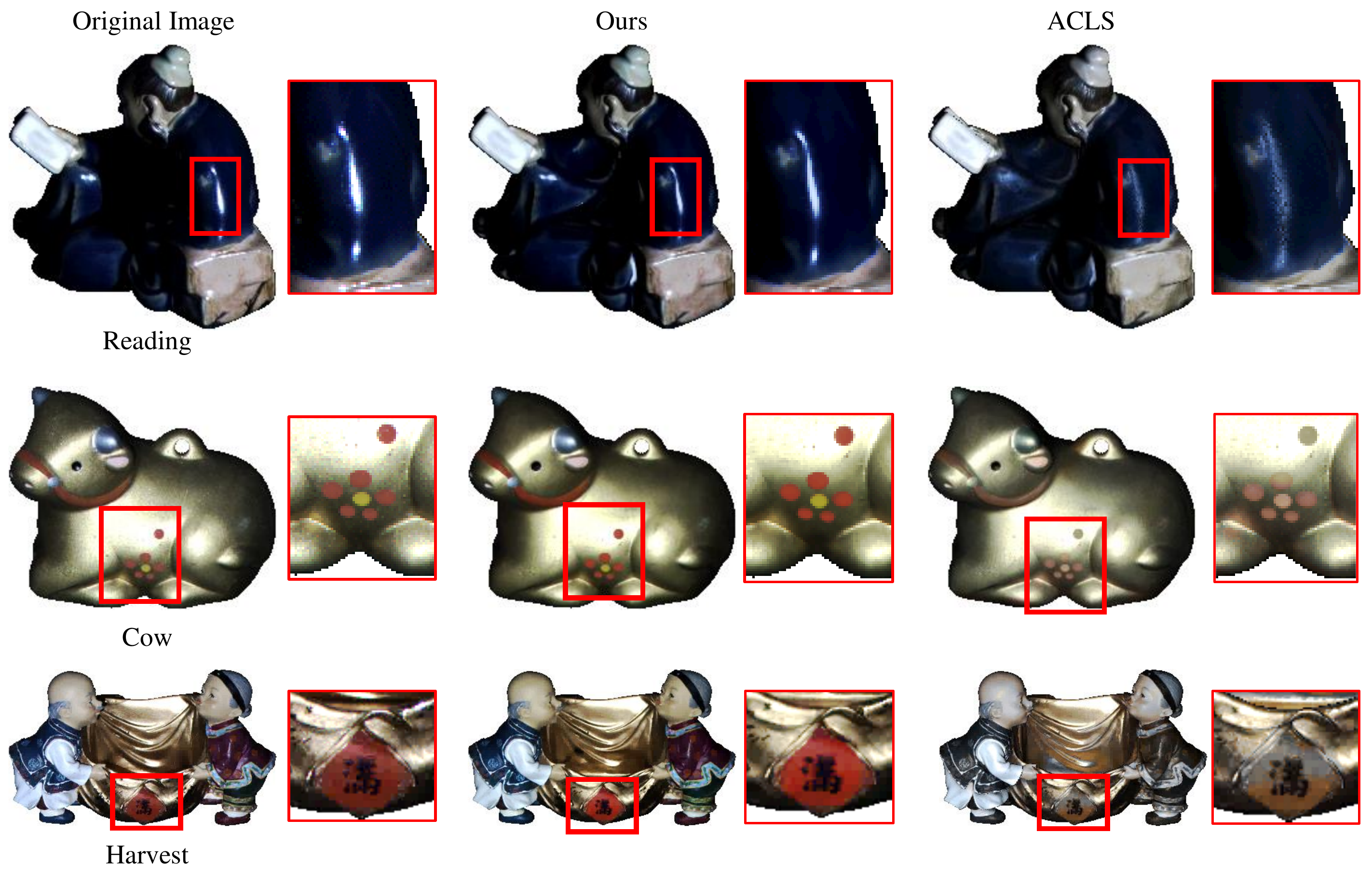}
\caption{Re-rendered image by our method and ACLS~\cite{alldrin2008photometric}. From left to right, we showcase the original image, the re-rendered image using our estimated neural svBRDFs, and the re-rendered image by ACLS~\cite{alldrin2008photometric}, respectively. Our method performs well on the specularities on ``Reading'', while ACLS failed to recover these highlights. Our method is also able to recover the spatially-varying materials, see the ``Cow'' and ``Harvest'', while ACLS failed to recover various materials on surfaces.
}
\label{fig:rendered_all}
\end{figure*}

\paragraph{Quantitatively evaluation} We re-rendered the observed image with our estimated reflectances, and ground truth lights. 
We compare our re-rendered images with ACLS~\cite{alldrin2008photometric}.
Note that ACLS's BRDF fitting results are provided by Shi~\etal~\cite{shi2016benchmark}, where the calibrated lightings, and the ground truth normal are used when fitting the BRDF. 
In comparison, our method only takes the calibrated lightings at the input.
The results are shown in \cref{tab:rerendered}. 
Our method achieves better reconstruction results on average. Our method performs particularly well on shiny objects and objects with variously materials, such as ``cow'', ``goblet'', ``harvest'', and ``reading''. Visual comparisons are shown in \cref{fig:rendered_all}. 

\begin{table*}[!h]
\centering
\caption{Quantitative comparison on the re-rendered images with our estimated BRDF. The metric here is peak signal-to-noise ratio (PSNR); higher is preferred. }
\small
\begin{tabular}{@{}lccccccccccc@{}}
\toprule
     & ball           & bear           & buddha         & cat            & cow            & goblet         & harvest        & pot1           & pot2           & reading        & Average        \\ \midrule
ACLS~\cite{alldrin2008photometric} & \textbf{40.27} & 46.75          & 40.27          & \textbf{46.87} & 44.35          & 42.67          & 33.90          & \textbf{51.55} & 52.06          & 30.60          & 42.93          \\
Ours & 37.82          & \textbf{47.96} & \textbf{41.14} & 46.31          & \textbf{45.85} & \textbf{43.56} & \textbf{36.24} & 51.40          & \textbf{52.36} & \textbf{31.58} & \textbf{43.42} \\ \bottomrule
\end{tabular}
\label{tab:rerendered}
\end{table*}


\newpage
\section{Material editing}
Our method opens up the possibilities to edit materials of objects. In the following figures, we select some surface points from the observed images and use the estimate reflectance (svBRDF) to re-render several new objects. 

\begin{figure*}[!h]
\centering
\includegraphics[width=0.9\textwidth]{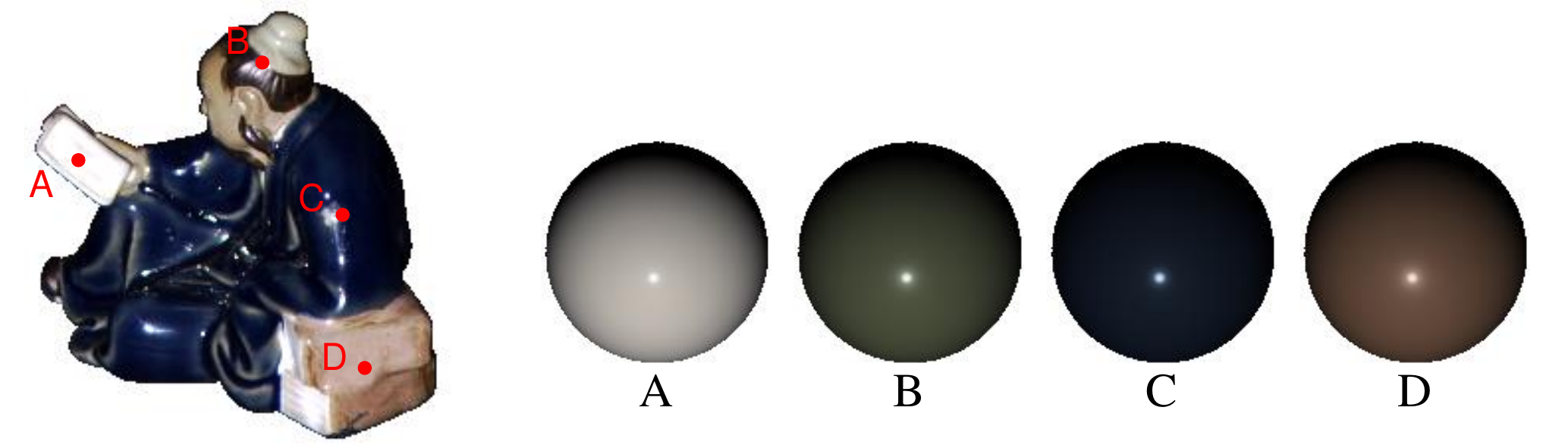}
\caption{\textbf{Visualization on the estimated svBRDFs.} This figure shows the estimated svBRDFs of the surface points on the objects. On the left is the observed image of ``Reading''. We select four different surface points on the object and showcase our re-rendered BRDF spheres of those points on the right side.
For better visualization, we normalize the BRDF spheres to have the maximum intensity to be $1$. The observed images are also scaled up for visualization.  
}
\label{fig:svbrdf_sphere}
\end{figure*}

\begin{figure*}[!h]
\centering
\includegraphics[width=\textwidth]{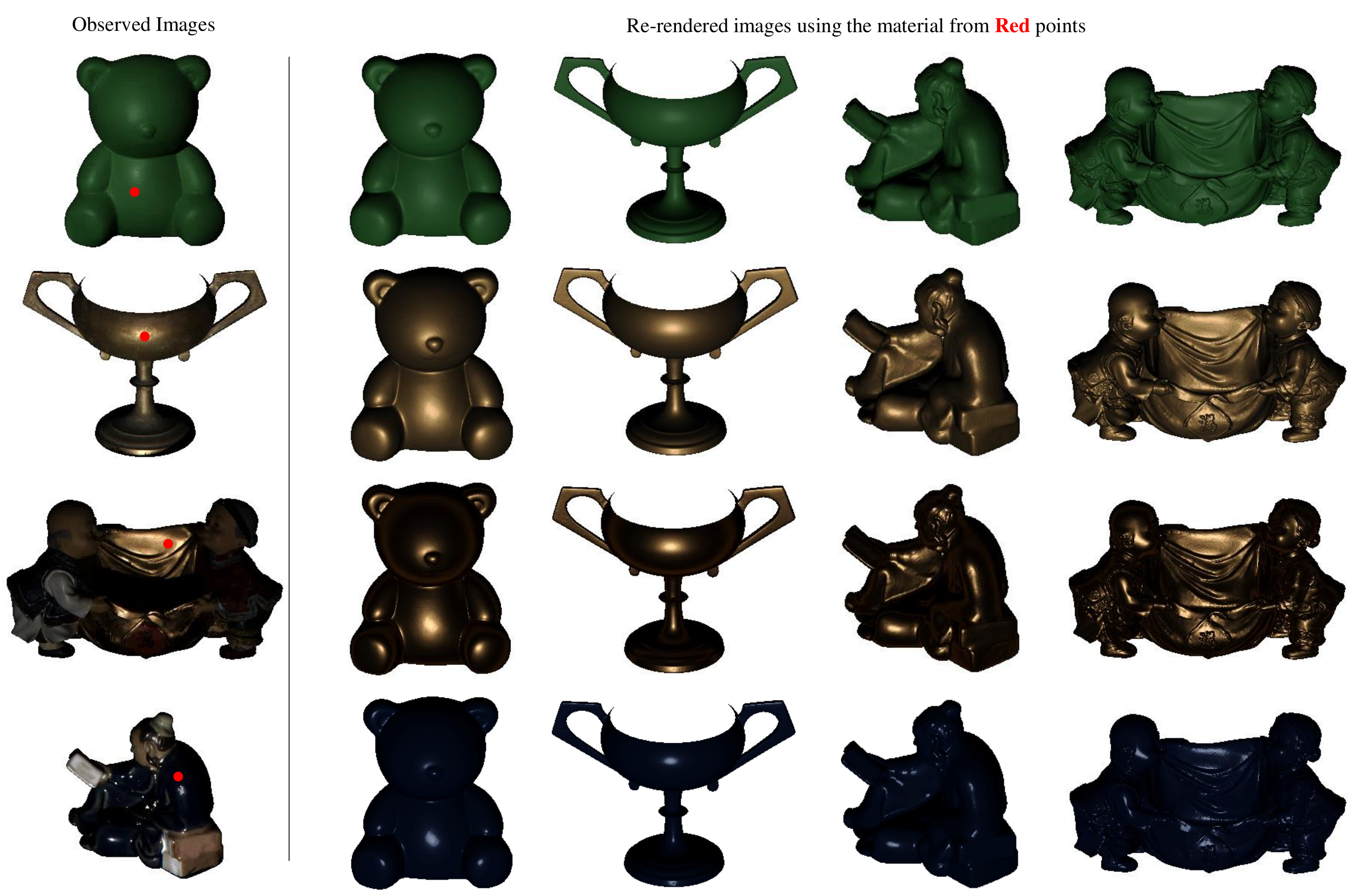}
\caption{\textbf{Material editing.} For each row, the left-most images are the observed images from DiLiGenT dataset. We use the estimated reflectance from the red points denoted in the left-most images to re-render several objects and present them on the right.
}
\label{fig:material_editing}
\end{figure*}

\section{BRDF evaluation on synthetic dataset.}
\begin{figure}[h]
\centering
\includegraphics[width=0.4\textwidth]{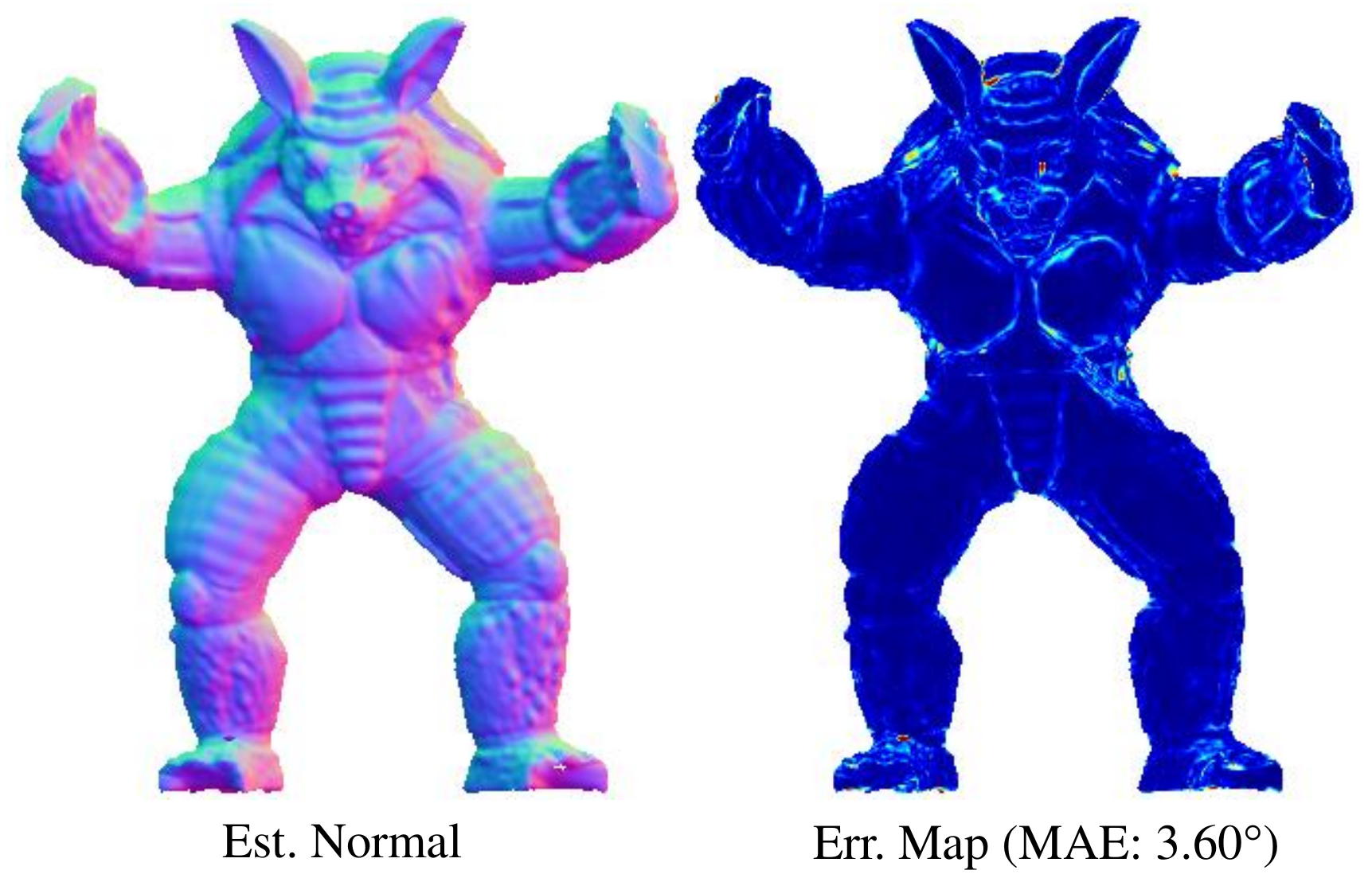}\includegraphics[width=0.4\textwidth]{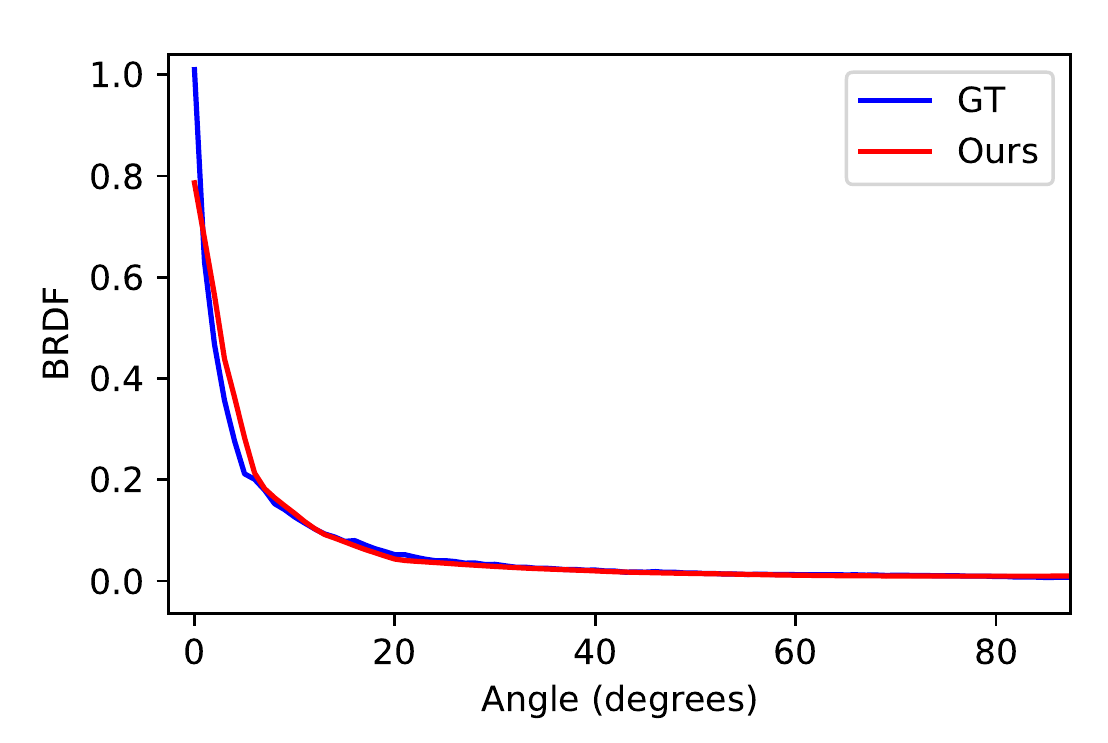}
\caption{From left to right, we showcase the estimated normal, error map, and our recovered slice of BRDF curve of ``Armadillo'' with ``alum-bronze'' as the material.}
\label{fig:sythn}
\end{figure}

In this section, we evaluate our method on a publicly available synthetic dataset\footnote{\url{https://github.com/guanyingc/UPS-GCNet}} proposed by Chen et al~\cite{chen2020learned}.
The dataset was rendered using the physically-based raytracer Mitsuba with MERL~\cite{Matusik:2003} as the BRDFs.
We showcase our results on ``Armadillo'' with ``alum-bronze'' as the material in \cref{fig:sythn}. From the result, even if the shape is as complicated as ``Armadilla'', our method can still recover the normals very well (with MAE $3.60^\circ$). We further plot a slice of our estimated BRDF curve and the ground truth BRDF curve in the right of \cref{fig:sythn}. We can see that our estimated BRDF is very close to the ground truth BRDF, which demonstrates that our method is robust in material recovery. 

\section{Additional Results}
\begin{figure*}
\centering
{\small
\hspace{1mm} GT Normal \hspace{15mm} Ours \hspace{23mm} TM18\cite{taniai2018neural} \hspace{17mm} PS-FCN\cite{chen2018ps} \hspace{15mm}  L2\cite{woodham1980photometric}}
\includegraphics[width=0.9\textwidth]{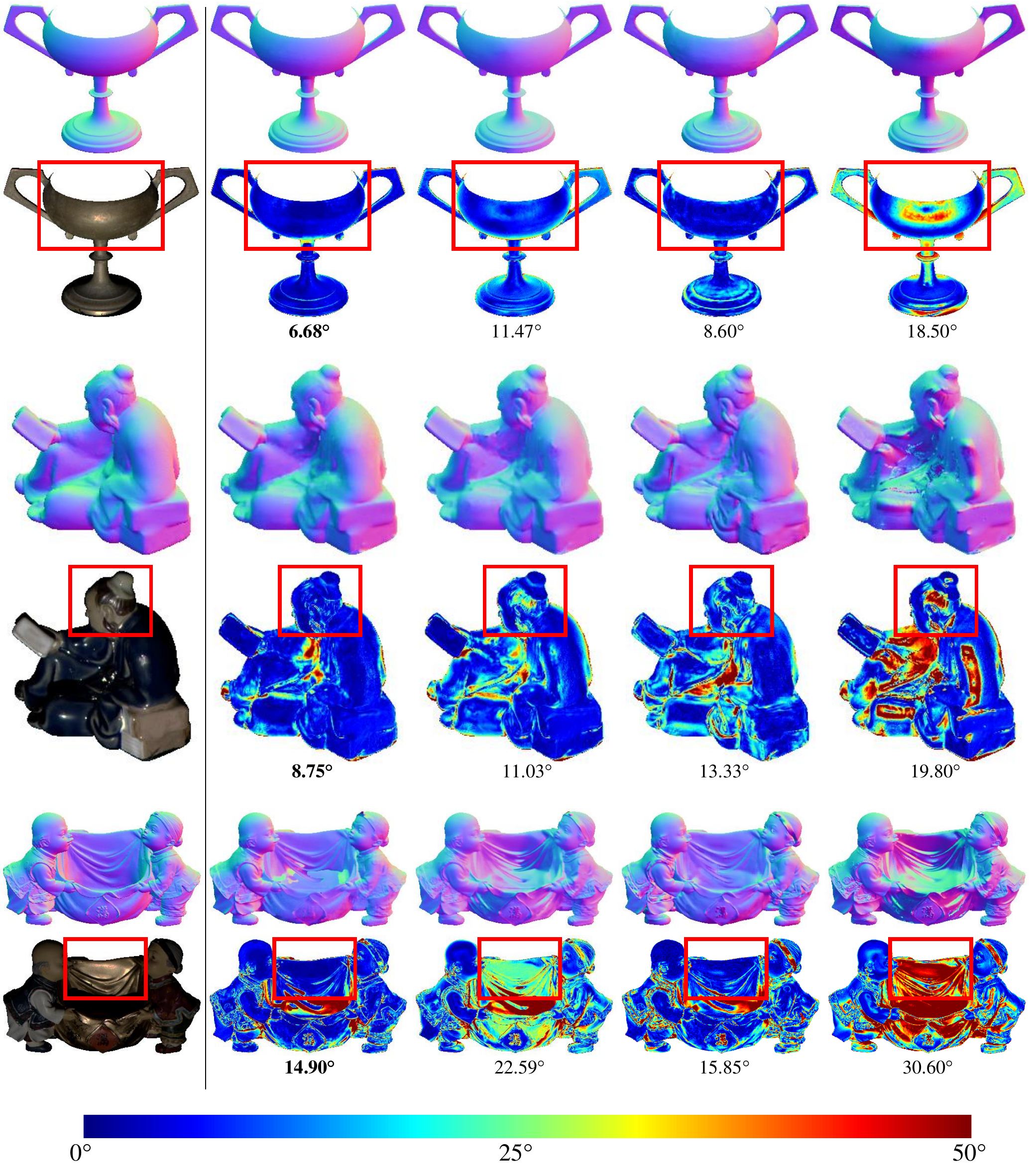}
\caption{\textbf{Normal estimation on specular objects: ``Goblet'', ``Reading'' and ``Harvest''.}
As shown in the observed image of these three objects, the ``Goblet'' is mostly made of metallic materials; ``Reading'' and ``Harvest'' present many specular effects over the clothes. 
Our method achieves the best performance in all these three objects, especially in those regions with high specularities. Please look at the red windows in the error map. 
``Reading'' contains many specularities over its cloth and its head. While all the other methods suffer on these specularities, our method still performs well in these regions, especially on the head.
The cloth of ``Harvest''  in the center also presents significant specular effects. While the other self-supervised method TM18~\cite{taniai2018neural} failed on these regions, our method correctly recover the surface normal.
These results demonstrate that our method is taking advantage of the information that the specularities provide. Hence, we can estimate the normal accurately on specular regions.}
\label{fig:normal_1}
\end{figure*}

\paragraph{Additional Results on Normal Estimation}
In \fref{fig:normal_1}, we present the normal estimation results on three specular objects: ``Goblet'', ``Reading'' and ``Harvest'' from DiLiGenT~\cite{shi2016benchmark}. These results demonstrate that our method is taking advantage of  the information that the specularities provide. Hence, we can estimate the normal accurately on specular regions.

\begin{sidewaysfigure*}
\centering
\includegraphics[width=0.99\textheight]{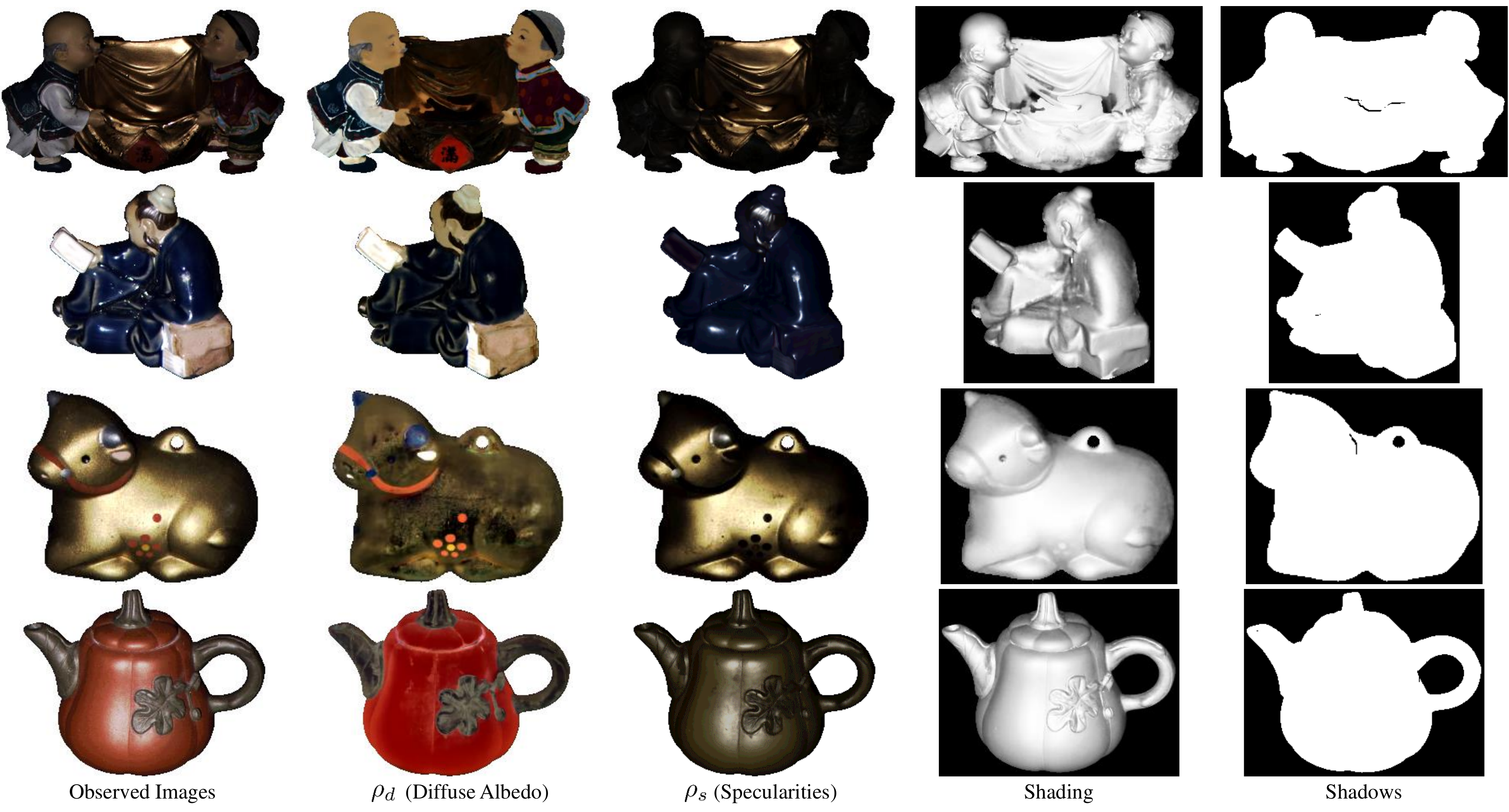}
\caption{\textbf{Visualization on each terms of the rendering equation}. 
In the above images, the first column displays the observed images of the objects.
The second and third column are the estimated diffuse albedo $\rho_d$, and specular components $\rho_s$. The fourth column is the shading map, which is computed by the dot product between light direction and surface normal ($\vl^T \vn)$. The last column is the estimated shadows, corresponding to $s$ in the equation. 
As seen from the diffuse albedo map in the cloth of the ``Harvest'' and the small patterns on the ``Cow'', the estimated diffuse albedo map retains the objects' fine details. These results demonstrate that our method can recover the fine details of the svBRDF map in the object. 
Note that, for better visualization, the images we selected here are all illuminated by a front light source. Hence, as can be seen in the observed images, there is little shadows. 
}
\label{fig:diffuse_spec}
\end{sidewaysfigure*}

\paragraph{Visualization on each terms of the rendering equation}
Recall the rendering equation (Eq.(1) in the main paper) is defined by
\begin{align}
    I = s \rho(\vl, \vv, \vn) \max(\vl^T \vn, 0),  \quad  \rho(\vl, \vv, \vn) = \rho_{\text{d}} + \rho_{\text{s}},
\end{align}
where $\rho_{\text{d}}$ is the diffuse albedo; $\rho_{\text{s}}$ is the specularities; $s$ is the shadows; $\max(\vl^T \vn, 0)$ is the shading term. 
In \fref{fig:diffuse_spec}, we present the visualization of our estimation on these terms.

\begin{figure*}
\centering
\includegraphics[width=0.7\textwidth]{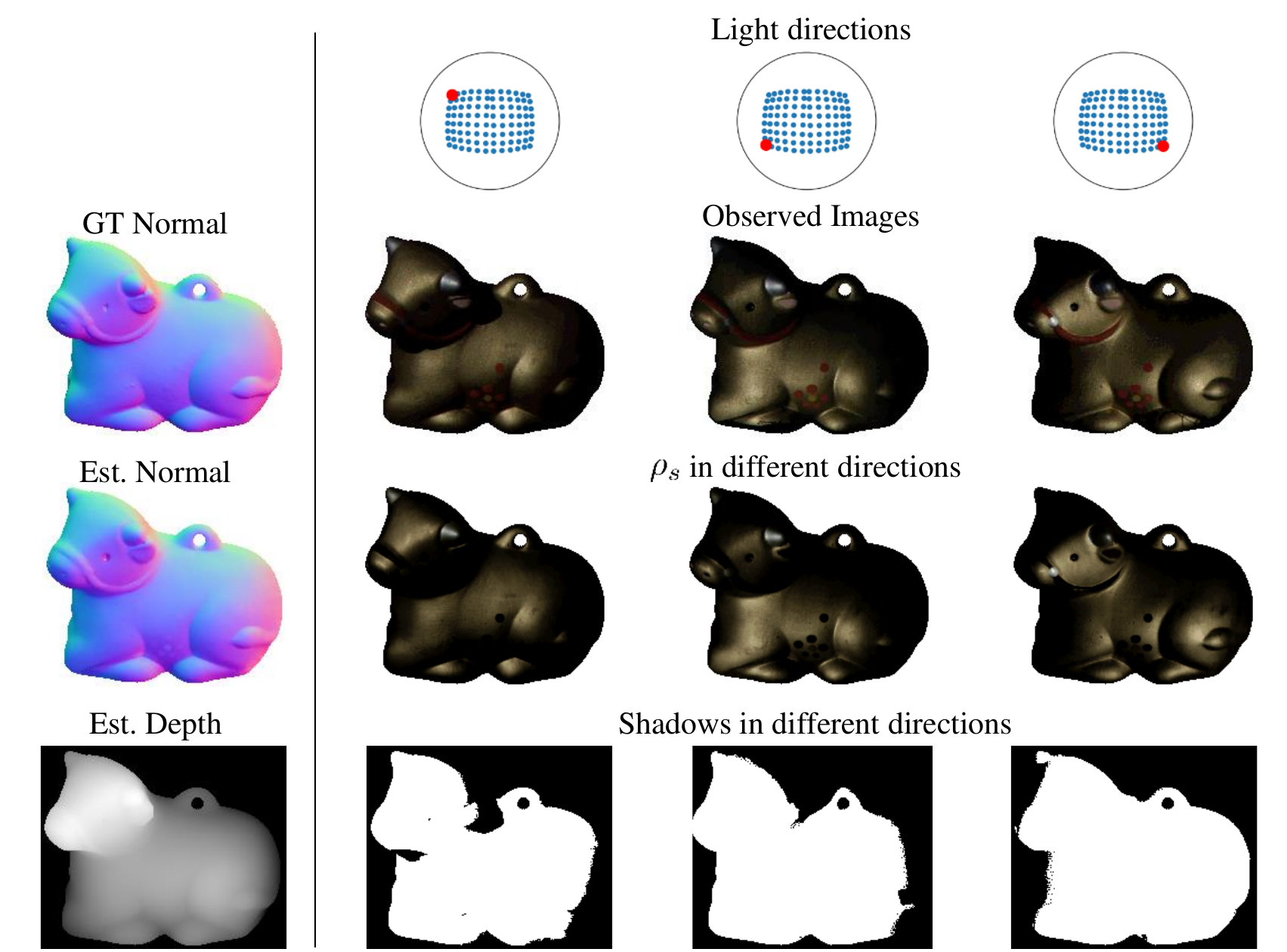}\\
(a) Cow\\
\includegraphics[width=0.7\textwidth]{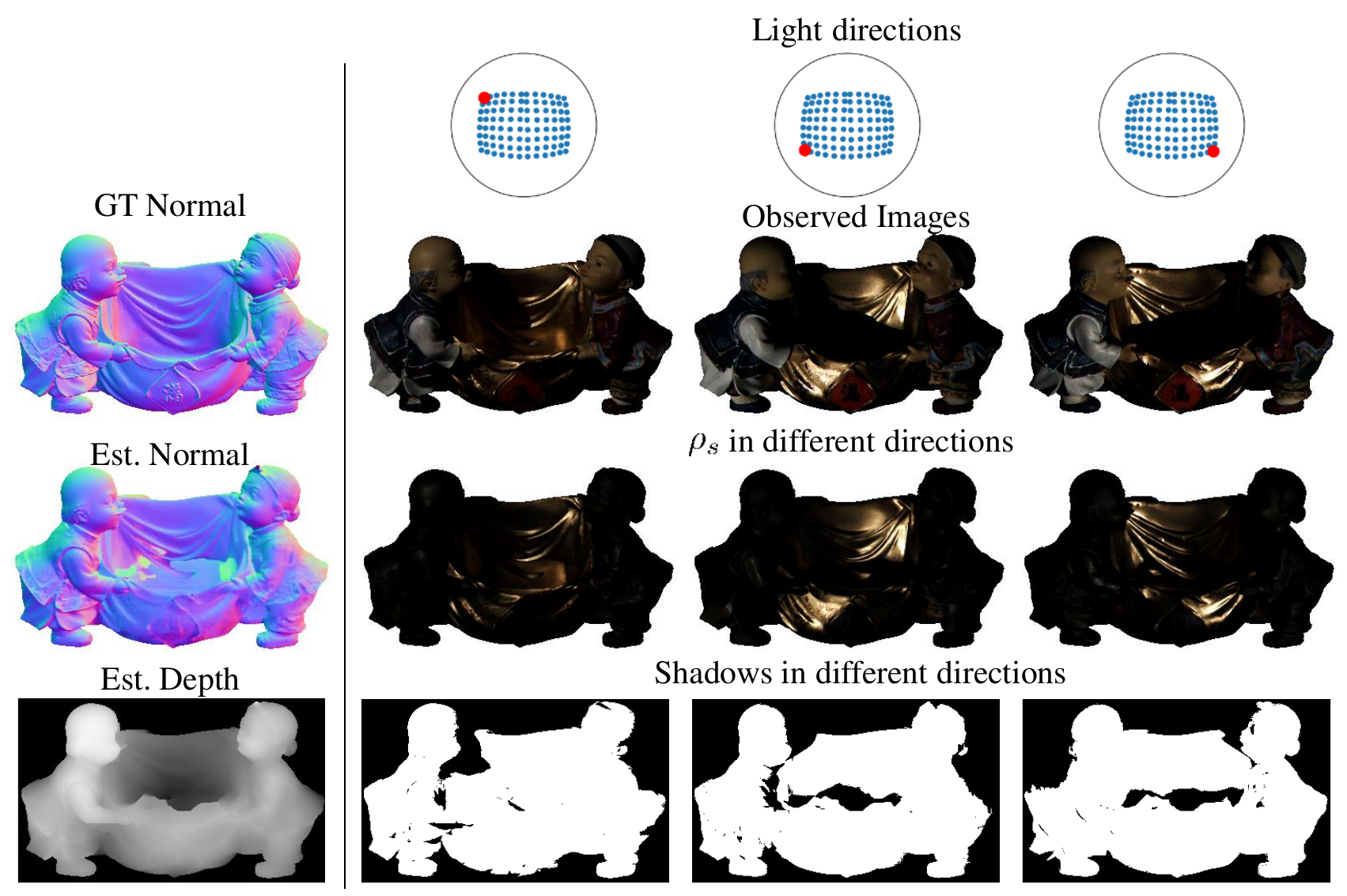}\\
(b) Harvest
\caption{\textbf{Estimated specularities and shadows under different illuminations.}
The leftmost column presents the ground truth normal and our estimated normal and depth as a reference of the object's geometry. We show the estimated specular components $\rho_s$ and estimated shadows under three different extreme lighting directions in the right-three columns.
In ``Cow'', the object is generally smooth, and our estimation of the shadows also visually match the observed images.
``Harvest'' has a complex geometry and consists of many depth discontinuities over the surface. As discussed in the Sec. 6 of the main paper, our method is influenced by these regions and will generate a ``shallower'' depth map. Hence, the estimated shadows are generally under-estimated. 
}
\label{fig:spec_shadow}
\end{figure*}

\paragraph{Additional Results on Shadows Estimation}
In \fref{fig:spec_shadow}, we showcase the estimated shadows and specularities under different light directions.

\begin{sidewaysfigure*}
\centering
\includegraphics[width= 0.9\textheight]{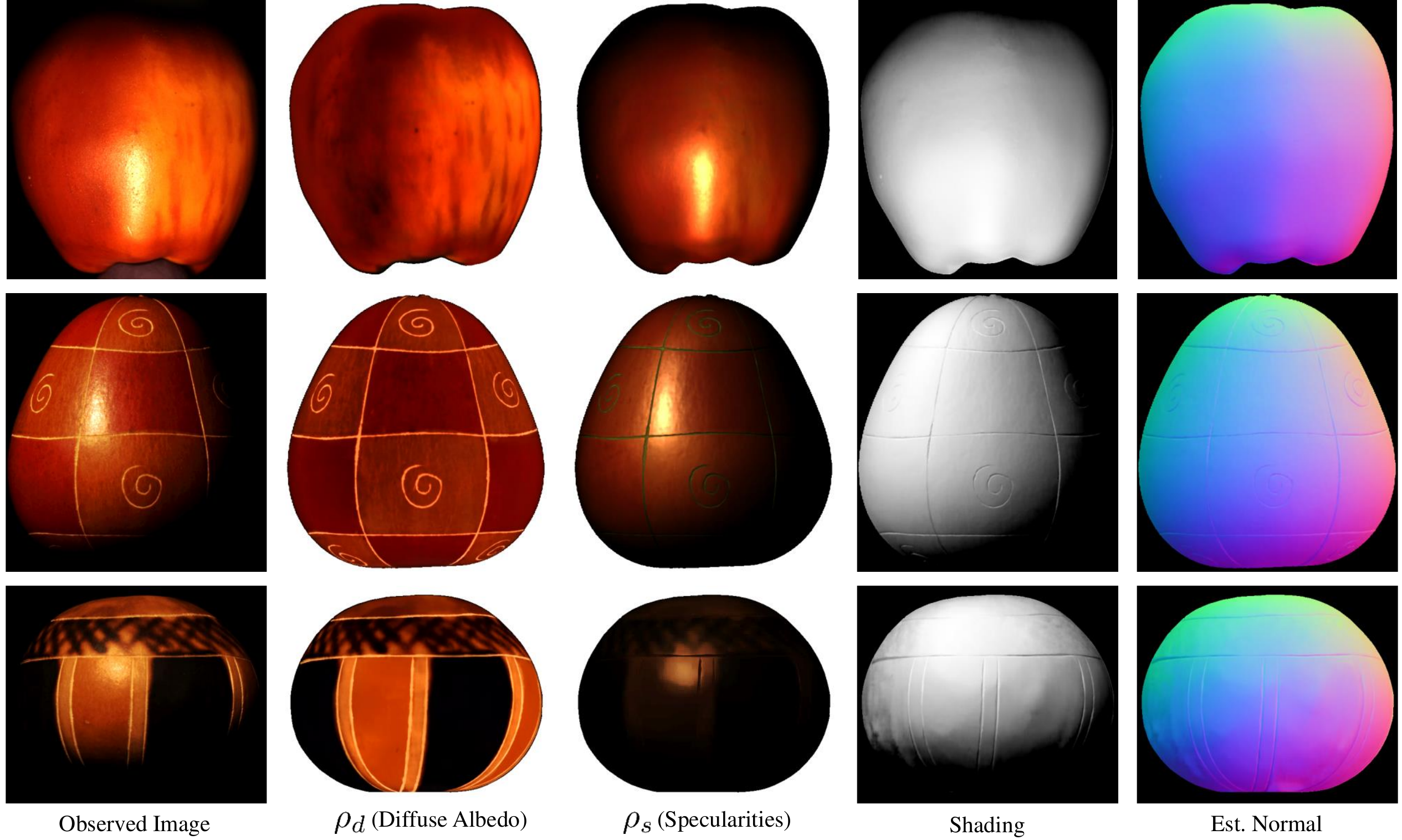}
\caption{\textbf{Results on Gourd\&Apple dataset}~\cite{alldrin2008photometric}. 
The columns from left to right are the observed images, our estimated diffuse albedo, specularities, shading, and surface normal of the objects.
}
\label{fig:apple_results}
\end{sidewaysfigure*}

\begin{figure*}
\centering
\includegraphics[width=\textwidth]{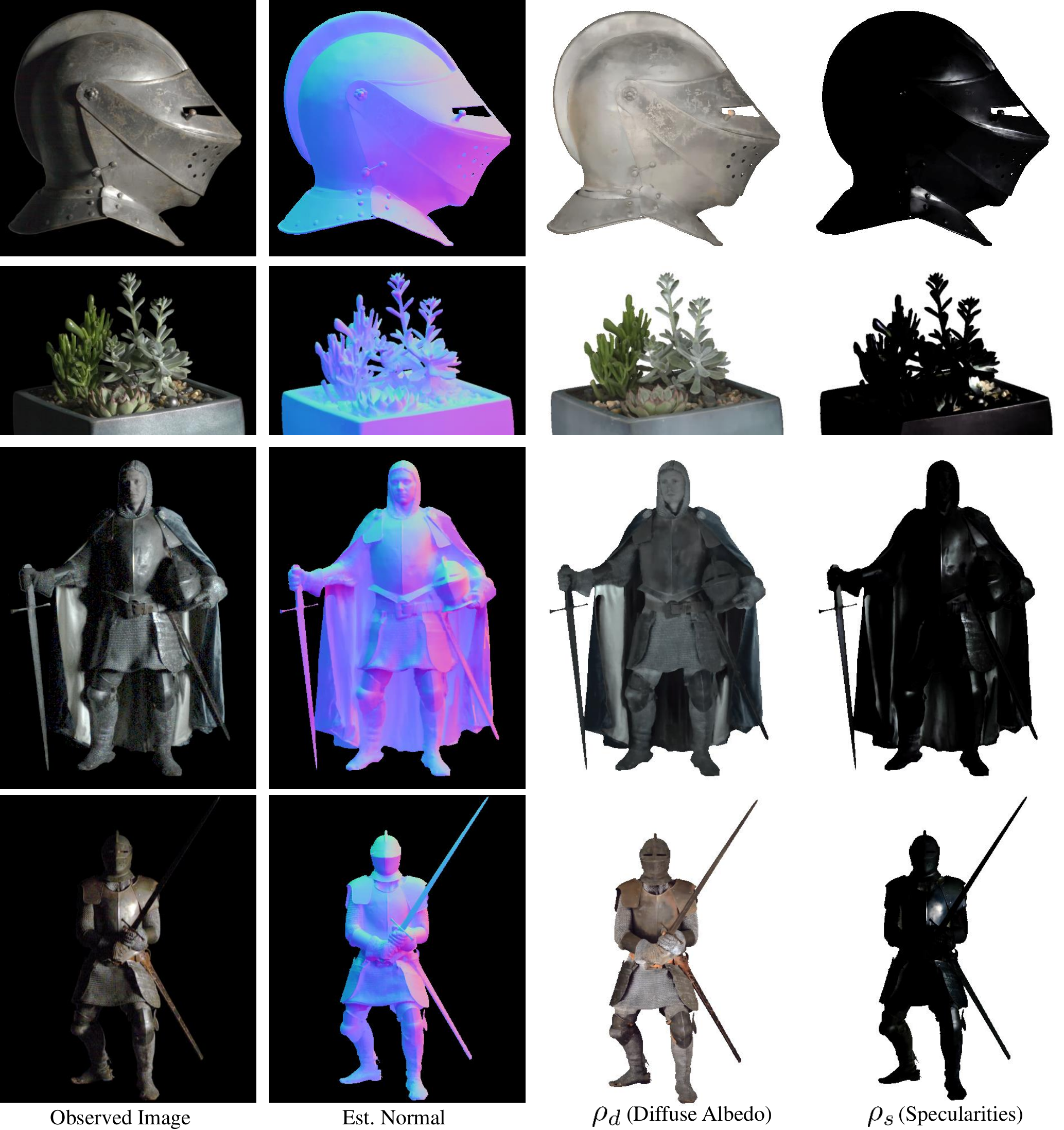}
\caption{\textbf{Results on Light Stage Data Gallery}~\cite{chabert2006relighting}.
The columns from left to right are the observed images, our estimated normal, diffuse albedo, and specularities of the objects.
}
\label{fig:lightstage_results}
\end{figure*}

\paragraph{Additional Results on Other Real-world Dataset}
We also test our method on two other challenging real-world datasets: Gourd\&Apple dataset~\cite{alldrin2008photometric} and  Light Stage Data Gallery~\cite{chabert2006relighting}, as shown in \fref{fig:apple_results} and \fref{fig:lightstage_results} separately.
Both of these two datasets do not provide ground truth normal for evaluation. Hence, we provide the visualization of the estimated normal, diffuse albedo, and specular map on these datasets. 
Our method correctly recovers the shape and materials of different objects. It also demonstrates that our method is robust on different objects with different materials.





\section{Ethics Statement}
With the advancement of photometric stereo, anyone can easily capture the 3D shape of a person's face.   The inverse rendering technique allows the user to alter the shape and appearance of an individual's face. The acquisition and alteration of such personal information, if without their consent, may lead to privacy and security breaching. Care must be taken to mitigate the potential risk of abusing this technique.

\end{document}